\documentclass[times, review, 10pt]{elsarticle}
\usepackage{stmaryrd}
\usepackage{bbm}
\usepackage{amsfonts}
\usepackage{amsmath}
\usepackage{mathrsfs}
\usepackage{amssymb}
\usepackage{graphicx}
\usepackage{subfigure}
\usepackage{caption}
\usepackage{booktabs}
\usepackage{graphicx}
\usepackage{setspace}
\usepackage[nodots]{numcompress}
\usepackage{algorithm}
\usepackage{algorithmic}
\usepackage{bm}
\usepackage{color}
\usepackage{float}
\usepackage{multirow}
\usepackage{geometry}
\usepackage{array}
\usepackage{url}
\usepackage{xcolor}
\usepackage{hyperref}

\begin{document}
\begin{frontmatter}
\title{Infrared-Assisted Single-Stage Framework for Joint Restoration and Fusion of Visible and Infrared Images under Hazy Conditions}
\author[]{Huafeng~Li$^{\scriptscriptstyle a,b}$}
\author[]{Jiaqi~Fang$^{\scriptscriptstyle a,b}$}
\author[]{Yafei Zhang$^{\scriptscriptstyle a,b}$\corref{mycorrespondingauthor}}
\author[]{Yu Liu$^{\scriptscriptstyle c}$}
\cortext[mycorrespondingauthor]{Corresponding author: E-mail:
zyfeimail@163.com}
%\thanks{~$^{1}$ indicates contributed to this work equally.}%co-first authors.
\address{a. Faculty of Information Engineering and Automation,
Kunming University of Science and Technology, Kunming 650500, Yunnan,
P.R. China.}
\address{b. Key Laboratory of Artificial Intelligence in Yunnan Province,
Kunming University of Science and Technology, Kunming 650500, Yunnan, P.R. China.}
\address{c. The School of Instrument Science and Opto-Electronic Engineering, Hefei University of Technology, Hefei  230009, China}

\begin{abstract}
Infrared and visible (IR-VIS) image fusion has gained significant attention for its broad application value. However, existing methods often neglect the complementary role of infrared image in restoring visible image features under hazy conditions. To address this, we propose a joint learning framework that utilizes infrared image for the restoration and fusion of hazy IR-VIS images. To mitigate the adverse effects of feature diversity between IR-VIS images, we introduce a prompt generation mechanism that regulates modality-specific feature incompatibility. This creates a prompt selection matrix from non-shared image information, followed by prompt embeddings generated from a prompt pool. These embeddings help generate candidate features for dehazing. We further design an infrared-assisted feature restoration mechanism that selects candidate features based on haze density, enabling simultaneous restoration and fusion within a single-stage framework. To enhance fusion quality, we construct a multi-stage prompt embedding fusion module that leverages feature supplementation from the prompt generation module. Our method effectively fuses IR-VIS images while removing haze, yielding clear, haze-free fusion results. In contrast to two-stage methods that dehaze and then fuse, our approach enables collaborative training in a single-stage framework, making the model relatively lightweight and suitable for practical deployment. Experimental results validate its effectiveness and demonstrate advantages over existing methods. The source code of the paper is available at \href{https://github.com/fangjiaqi0909/IASSF}{\textcolor{blue}{https://github.com/fangjiaqi0909/IASSF}}.

\end{abstract}
\end{frontmatter}
\section{Introduction}
Infrared and visible (IR-VIS) image fusion effectively combines the unique information from both infrared and visible images, creating a composite image that integrates their complementary features. This fused image not only provides a comprehensive and accurate scene representation but also significantly aids observers in understanding and analyzing complex environments. Consequently, this technology holds tremendous potential and value in fields such as military reconnaissance, aerospace, environmental monitoring, and medical diagnostics.

\begin{figure}[htbp]
	\centering
	\includegraphics[width=0.6\textwidth]{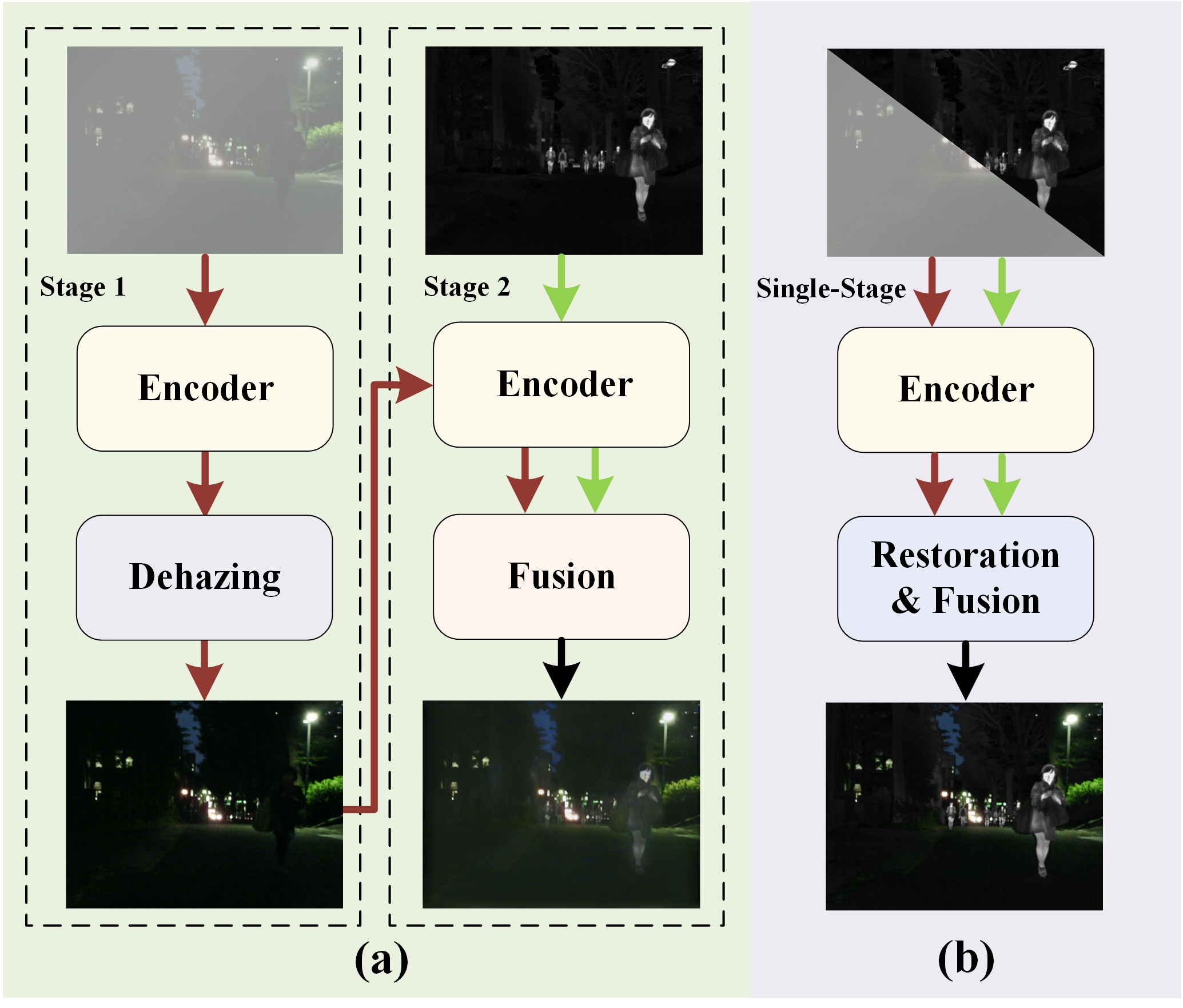}
	\caption{Comparison of existing method and our method for hazy IR-VIS
		image fusion. (a) The existing method, (b) Our method.}
	\label{Fig1}
\end{figure}

In recent years, the emergence of deep learning has rapidly advanced numerous areas within computer vision\cite{57,58,60,62}, and infrared–visible (IR–VIS) image fusion has achieved significant progress\cite{54,cdtfusion}; however, existing methods generally assume that the input visible images are of good visual quality. In hazy conditions, visible images are affected by haze, resulting in unclear imagery, which makes it difficult for these methods to generate clear, haze-free fusion results. Traditional approaches typically address this issue by first applying a dehazing algorithm to the hazy image and then fusing the dehazed image with the infrared image, as shown in Fig. \ref{Fig1}(a). Although this two-stage strategy is feasible, it fails to integrate the dehazing and fusion tasks into a unified framework for joint training, making it challenging to balance the relationship between the two tasks. While dehazed images may show good dehazing performance, they are not always optimal for subsequent fusion tasks. Additionally, the two-stage process of dehazing followed by fusion involves different methodologies, reducing the model's compactness.

To address the issues arising from the two-stage processing paradigm, Li et al. \cite{30} proposed the all-weather multi-modality image fusion method, which achieves image restoration and fusion under various complex weather conditions. However, this method fails to effectively coordinate the differences between the different restoration tasks, limiting further improvements in restoration and fusion performance. In response, Yi et al. \cite{31} introduced a method called Text-IF, which guides the fusion of degraded images using semantic text. Nevertheless, this approach relies on pre-input text descriptions, increasing the complexity of model deployment. Furthermore, while Text-IF is designed for the restoration and fusion of multiple types of degraded images, it faces challenges in balancing fusion performance across various degradation scenarios without compromising individual task performance.

In response to the challenge of IR-VIS image fusion and restoration under hazy conditions, we propose an infrared-assisted joint learning framework, as shown in Fig. \ref{Fig1}(b). To mitigate the impact of discrepancies between IR-VIS images on hazy image feature restoration, we design a prompt generation mechanism. It leverages non-shared information from input images to create a prompt selection matrix that selects and generates prompt embeddings from a prompt pool. These embeddings act as candidate features to aid in the recovery of hazy image features. For effective restoration of haze-affected features, we construct an infrared-assisted feature restoration module. It guides the selection of candidate features based on haze density to restore visible image features impacted by haze, enabling the joint processing of restoration and fusion within a single-stage framework. In this process, our focus shifts from solely enhancing the restoration of hazy visible images to emphasizing how restored features can further improve the quality of the fusion results.

To further enhance the fusion effect, we propose a multi-stage prompt embedding fusion module, which strengthens feature restoration and fusion with the help of the feature supplementation capability of the prompt generation. The proposed method not only effectively fuses IR-VIS images but also eliminates the interference of haze, producing clear and haze-free fusion results. Compared to the traditional two-stage approach of first dehazing and then fusing, our method more fully exploits the correlation between dehazing and fusion tasks, achieving a balance between them through a single-stage framework with collaborative training. Furthermore, the model structure is relatively lightweight and compact, facilitating practical deployment. Unlike existing multi-task fusion frameworks, our method is specifically designed for IR-VIS image fusion and restoration under hazy conditions, demonstrating excellent fusion and restoration performance. Therefore, our approach enriches the technical system for IR-VIS image fusion under hazy conditions and provides a new perspective for the restoration and fusion of low-quality images. In summary, the main contributions and advantages of our method are reflected in the following aspects:

\begin{itemize}
\item We propose an infrared-assisted single-stage framework for IR–VIS fusion under hazy conditions, jointly optimizing dehazing and fusion. Compared with two-stage pipelines, it better exploits their correlation to achieve more balanced results. The compact design efficiently coordinates both processes, enabling restored features to produce clearer, haze-free, high-quality fused images and improving practical applicability.

\item We design a prompt generation module that leverages non-shared information to construct a prompt selection matrix, enabling adaptive prompt selection and embedding to assist visible image restoration in hazy scenarios and alleviate modality discrepancies between infrared and visible features. In addition, we propose a multi-layer fusion mechanism based on prompt embeddings for feature compensation and progressive refinement, further improving the fusion quality.

\item Extensive experiments conducted on the MSRS, M3FD, and RoadScene datasets demonstrate that the proposed framework, while maintaining relatively low model complexity, achieves superior or comparable performance to a wide range of recent fusion and dehazing methods in terms of both subjective visual quality and objective evaluation metrics.
\end{itemize}

\section{Related Work}\label{section:related work}
\subsection{Typical Fusion Methods}
In IR-VIS image fusion, traditional methods based on multi-scale transforms and sparse representations \cite{14,13,17,1} remain relevant. However, deep learning-based techniques have become mainstream. These methods can be categorized into three types: Convolutional Neural Network (CNN)-based methods \cite{2,4,5}, hybrid CNN-Transformer methods \cite{3,20,21,22}, and Generative Adversarial Network (GAN)-based methods \cite{12,18,19,16,11}. CNN-based methods extract features from input images and perform fusion using specialized modules, enhancing image details and contrast. However, CNNs are limited in modeling long-range dependencies, which impacts fusion quality in complex scenes. In contrast, Transformers excel at capturing long-range dependencies but struggle with local details and edges. Hybrid methods, such as AFT \cite{20}, YDTR \cite{21}, and HitFusion \cite{22}, combine CNN with Transformer to model local and global information, improving fusion performance.

In GAN-based methods, FusionGAN \cite{12} uses a single discriminator to fuse IR-VIS images, which does not maintain modality balance, leading to biased fusion results. To address this, subsequent research introduced dual discriminator-based GAN methods. For example, LGMGAN \cite{18} combines a Conditional GAN with dual discriminators to fuse multi-modality information effectively. DDcGAN \cite{19} uses dual discriminators for multi-resolution fusion, improving consistency across scales. Moreover, AttentionFGAN \cite{11} integrates an attention mechanism to focus on important feature regions, significantly enhancing fusion performance. However, these methods assume that the images to be fused are of high quality, which makes it challenging to produce high-quality fusion results under hazy conditions.
\subsection{Methods Under Complex Imaging Conditions}

Under complex imaging conditions, various factors affect the quality of visible images. Thus, achieving high-quality fusion results under these conditions has become a crucial research direction in the field of IR-VIS image fusion. In low-light conditions, PIAFusion \cite{36} improves IR-VIS image fusion by introducing an illumination-aware loss function. DIVFusion \cite{24} enhances dark areas, details, and reduces color distortion by separating scene illumination and enhancing texture contrast, achieving high-quality fusion in nighttime conditions. IAIFNet \cite{25} uses an illumination enhancement network along with adaptive difference fusion and salient object awareness modules to better fuse features in IR-VIS images. LENFusion \cite{26} generates high-contrast fusion results through three stages: brightness adjustment, enhancement, and feedback. For low-resolution images, HKD-FS \cite{27} employs knowledge distillation to convert low-resolution IR-VIS images into high-resolution outputs. MLFusion \cite{28} incorporates meta-learning into the IR-VIS image fusion framework, enabling fusion from inputs of any resolution to outputs of any resolution.

To address the degradation of visible images under complex conditions, a decomposition-based and interference-aware fusion method was proposed in \cite{29}, which is capable of handling multiple degradations such as noise, overexposure, and snow, but does not involve hazy scenarios. To tackle haze, AWFusion \cite{30} introduces a clear feature prediction module based on the atmospheric scattering model, thereby enabling dehazing capability. However, AWFusion simultaneously considers various weather conditions such as snow and rain, which reduces its effectiveness specifically in hazy scenarios. To balance multiple tasks, Text-IF \cite{31} employs text guidance and generates modulation parameters to control cross-modal attention outputs. Nevertheless, Text-IF is not specifically designed for hazy IR–VIS fusion, thus showing limited performance in such conditions, and the requirement of textual input also limits its practicality. OmniFuse \cite{OmniFuse} and Text-DiFuse \cite{TextDiFuse} explicitly couple diffusion models with multimodal fusion, removing compound degradations and integrating information either in the latent space or during the diffusion process, while achieving controllable enhancement of specific semantic targets via text modulation. However, these approaches mainly focus on general compound degradations and lack targeted modeling for hazy IR–VIS fusion, while their dependence on textual or detection modules introduces additional cost for real deployment. Deno-IF \cite{DenoIF} addresses multi-noise infrared–visible fusion and obtains high-quality results from noisy inputs through unsupervised denoising and feature restoration, but mainly focuses on noise degradation rather than hazy conditions. In addition, CFMW [35] achieves joint optimization of multi-weather removal and visible–infrared object detection through a weather-removal diffusion model and a cross-modal fusion Mamba architecture. However, CFMW mainly targets the detection task rather than generating high-quality hazy fusion images. VIFNet \cite{32} restores hazy images using infrared guidance, but focuses only on dehazing and does not perform fusion. In contrast, this paper specifically targets IR–VIS fusion under hazy conditions and aims to obtain clear, haze-free fusion results.

\section{Proposed Method}\label{section:proposed_method}
\subsection{Overview}

\begin{figure*}[t!]
	\centering
	\includegraphics[width=1.0\textwidth]{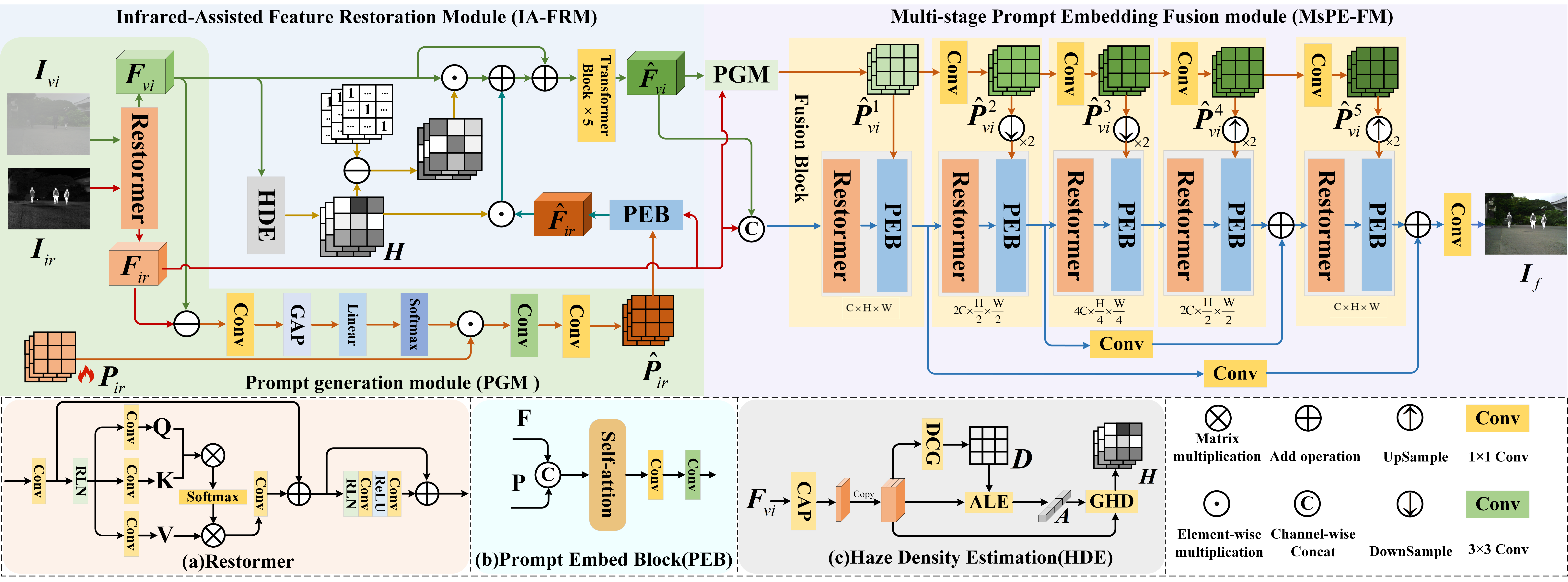}\vspace{-2mm}
	\caption{Overall framework of the proposed method. The input IR and hazy VIS image pair $\left\{{{\bm{I}}_{ir}}, {{\bm{I}}_{vi}} \right\}$ is processed by the PGM to obtain features \(\left\{ {{\bm{F}}_{ir}},{{\bm{F}}_{vi}} \right\}\) and a prompt \({{\bm{\hat P}}_{ir}}\) for \({{\bm{F}}_{ir}}\). Through the PEB, the prompt embedding \({{\bm{\hat P}}_{ir}}\) is used to refine the IR feature ${{\bm{F}}_{ir}}$, reducing redundant information and generating the refined IR feature ${{\hat{\bm{F}}_{ir}}}$. The haze density esitimation (HDE) module \cite{34} estimates the haze density in the VIS features to dynamically adjust the proportion of injected IR information, preventing excessive IR injection. The Transformer block removes degradation from the input features to obtain haze-free features. In the MsPE-FM, the haze-free VIS features and IR features are combined and passed to the Fusion Block for feature fusion. The PGM and PEB further are used to enhance the IR-VIS complementary information, reconstructing the final fused image.}\vspace{-2mm}
	\label{Fig2}
\end{figure*}

As shown in Fig. \ref{Fig2}, our method comprises three core modules: the Infrared-Assisted Feature Restoration Module (IA-FRM), the Prompt Generation Module (PGM), and the Multi-stage Prompt Embedding Fusion Module (MsPE-FM). IA-FRM leverages infrared image features to assist in restoring lost information in heavily hazy regions of visible images, making it easier to restore these hazy areas. PGM generates a set of prompts to overcome the limitations of infrared images when assisting in the restoration of features in these dense hazy regions. Using the restored visible image features and prompts from PGM, MsPE-FM performs the fusion of IR-VIS image features, reconstructing a haze-free fused result. 

\subsection{Prompt Generation}
Infrared imaging sensors maintain performance in hazy conditions, allowing them to penetrate heavy haze. In a rigorously registered pair of IR-VIS images $\{\bm{I}_{ir}, \bm{I}_{vi}\}$ provided to the model, we assume that only the visible image $\bm{I}_{vi}$ contains haze, while the infrared image $\bm{I}_{ir}$ is unaffected by haze. The core challenge of this work is to effectively utilize the infrared image $\bm{I}_{ir}$ to restore the visible image $\bm{I}_{vi}$ and then fuse them. However, the significant modality differences between IR-VIS images make it difficult to rely solely on the infrared image $\bm{I}_{ir}$ to recover the details lost in the hazy visible image. To overcome these challenges, this paper proposes the PGM, which generates a prompt embedding to create compensatory features that address the limitations of infrared features.

As shown in Fig.~\ref{Fig2}, in the PGM, we first utilize an encoder constructed with Restormer~\cite{33} to perform feature encoding on the input registered IR-VIS images $\{\bm{I}_{ir}, \bm{I}_{vi}\}$. As depicted in Fig.~\ref{Fig2}(a), Restormer consists of a self-attention layer and a feed-forward network layer. The features output by the Restormer encoder are denoted as $\bm{F}_{vi} \in \mathbb{R}^{C \times H \times W}$ and $\bm{F}_{ir} \in \mathbb{R}^{C \times H \times W}$, where $C$, $H$, and $W$ represent the number of channels, height, and width of the features, respectively. Additionally, in this module, the features of the hazy visible image $\bm{I}_{vi}$ and the infrared image $\bm{I}_{ir}$ are processed by
\begin{equation}
	\bm{F}_{vi-ir} = \bm{F}_{vi} - \bm{F}_{ir}
\end{equation}
to remove the shared information and highlight the unique information. The resulting difference $\bm{F}_{vi-ir}$ is then fed into a weight prediction network composed of Convolutional (Conv) layer, GAP, Linear layer, and Softmax, resulting in a weight matrix $\bm{W}_p \in \mathbb{R}^{C \times H \times W}$ for selecting prompt information from the prompt pool $\bm{P}_{ir}$. We implement $\bm{P}_{ir}$ as a set of learnable prompts rather than a single fixed feature map:
	\begin{equation}
		\bm{P}_{ir} = \big\{\mathcal{P}_{ir}^{l}\big\}_{l=1}^{L}, 
	\end{equation}
	Each $\mathcal{P}_{ir}^{l}$ is a learnable prompt. During training, the $L$ prompts are randomly initialized and jointly optimized together with the backbone network. Subsequently, these prompts are stacked into a learnable tensor, resulting in the prompt representation $\bm{P}_{ir} \in \mathbb{R}^{C \times H \times W}$.

At this stage, the prompt embedding generated for compensating the infrared image features can be represented as
\begin{equation}
	\hat{\bm{P}}_{ir} = \mathrm{Conv}_{3 \times 3} \left( \mathrm{Conv}_{1 \times 1} \left( \bm{W}_p \odot \bm{P}_{ir} \right) \right),
\end{equation}
where $\mathrm{Conv}_{3 \times 3}$ and $\mathrm{Conv}_{1 \times 1}$ denote $3\times3$ and $1\times1$ Conv layers, respectively. The spatially varying weight matrix $\bm{W}_p$ predicted from $\bm{F}_{vi-ir}$ modulates the prompt tensor via element-wise multiplication, producing a content-adaptive prompt representation that is further refined by the subsequent convolutional layers to obtain $\hat{\bm{P}}_{ir}$. The resulting $\hat{\bm{P}}_{ir}$ is then fed, along with the infrared image features $\bm{F}_{ir}$, into the Prompt Embedding Block (PEB) to obtain the features $\hat{\bm{F}}_{ir}$ for compensating the dense haze regions in the visible image $\bm{I}_{vi}$.

\subsection{Feature Restoration Assisted by Infrared Image}

To effectively utilize the information provided by the infrared image $\bm{I}_{ir}$ for restoring features in haze regions, we design the IA-FRM. As shown in Fig.~\ref{Fig2}, when restoring features in hazy images, regions with higher haze density should receive more focus. Therefore, it is essential to estimate the haze density in the input images. To achieve this, we adopt the method from \cite{34} to estimate the haze density in the visible branch, as illustrated in Fig.~\ref{Fig2}(c).

Let $\bm{F}_{vi} \in \mathbb{R}^{C \times H \times W}$ denote the visible feature map output by the encoder. In the HDE, we first apply Channel Average Pooling (CAP) to $\bm{F}_{vi}$ to obtain a single-channel feature map
\begin{equation}
	\bar{\bm{F}}_{vi} = \mathrm{CAP}(\bm{F}_{vi}) \in \mathbb{R}^{1 \times H \times W},
\end{equation}
and then replicate it along the channel dimension (the “Copy” operation in Fig.~\ref{Fig2}(c)) to form a pseudo-RGB feature $\tilde{\bm{F}}_{vi} \in \mathbb{R}^{C \times H \times W}$.
Based on $\tilde{\bm{F}}_{vi}$, the Dark Channel Generation (DCG) module computes the dark channel
\begin{equation}
	D(i,j) = \min_{c \in \{1,2,3\}} \left( \min_{(u,v)\in \Omega(i,j)} \bm{\tilde{F}}_{vi}^{c}(u,v) \right),
\end{equation}
where $\Omega(i,j)$ denotes a local window centered at pixel $(i,j)$ and $\bm{\tilde{F}}_{vi}^{c}$ is the $c$-th channel of $\tilde{\bm{F}}_{vi}$. The dark channel $D$ highlights regions that are heavily affected by haze.

\begin{figure*}[t!]
	\centering
	\includegraphics[width=1.0\textwidth]{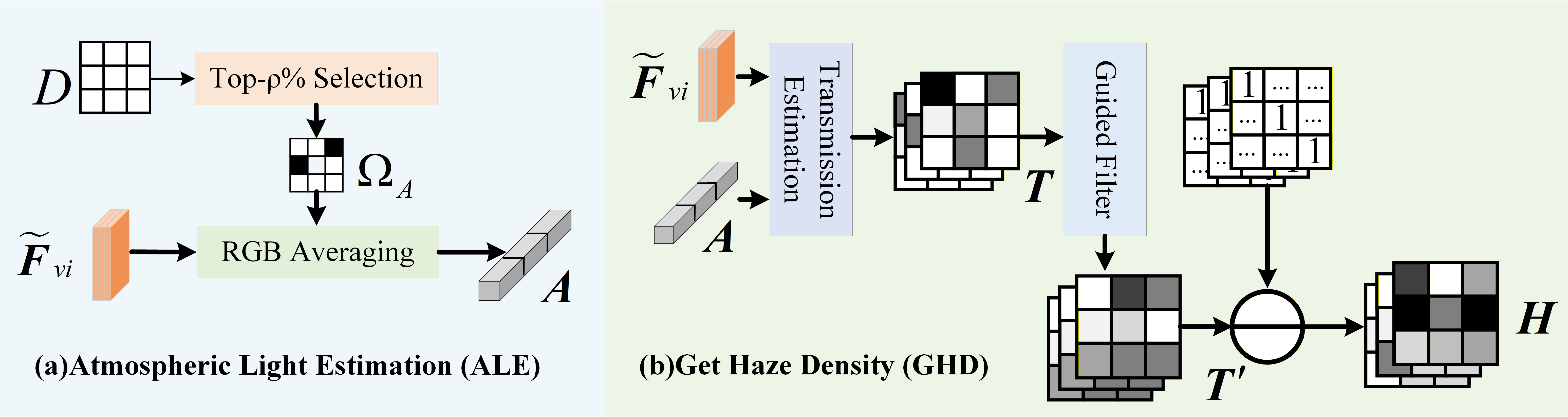}\vspace{-2mm}
	\caption{(a)Illustration of the Atmospheric Light Estimation (ALE) module.(b)Illustration of the Get Haze Density (GHD) module.}\vspace{-2mm}
	\label{ALE+GHD}
\end{figure*}

In the Atmospheric Light Estimation (ALE) module,  As shown in Fig.~\ref{ALE+GHD}(a),we first select the top $\rho\%$ brightest pixels in $D$ (set to $\rho = 0.1$ as suggested in \cite{34}) and denote their index set as $\Omega_A$. The global atmospheric light $\bm{A} \in \mathbb{R}^{3}$ is then estimated by averaging the corresponding intensities of $\tilde{\bm{F}}_{vi}$:
\begin{equation}
	\bm{A} = \frac{1}{|\Omega_A|} \sum_{(i,j)\in \Omega_A} \tilde{\bm{F}}_{vi}(i,j).
\end{equation}

In the Get Haze Density (GHD) block,  As shown in Fig.~\ref{ALE+GHD}(b), the estimated atmospheric light $\bm{A}$ is first broadcast along the channel dimension and used to normalize the visible features. By combining $\bm{A}$ with $\tilde{\bm{F}}_{vi}$, we obtain the initial transmission map
\begin{equation}
	\bm{T} = \mathbf{1} - \omega \cdot \mathrm{DCG}\!\left( \tilde{\bm{F}}_{vi} \odot \bm{A}^{-1} \right),
\end{equation}
where $\omega$ is a constant that adjusts the effect of the DCG prediction (set to $0.95$ as suggested in \cite{34}); $\mathbf{1} \in \mathbb{R}^{1 \times H \times W}$ is a matrix of ones; $\bm{A}^{-1}$ represents the element-wise reciprocal of $\bm{A}$; and $\odot$ denotes the Hadamard product.

Next, the coarse transmission map $\bm{T}$ is refined using a guided filter, yielding the refined transmission map $\bm{T}'$. Since the transmission map is inversely proportional to the haze degree, the haze density map $\bm{H}$ is estimated as
\begin{equation}
	\bm{H} = \mathbf{1} - \bm{T}'.
\end{equation}
Pixels with higher values in $\bm{H}$ correspond to regions with heavier haze. Therefore, $\bm{H}$ is used to select the information for restoring $\bm{F}_{vi}$ from $\hat{\bm{F}}_{ir}$. In this process, we use $\left( \mathbf{1} - \bm{H} \right)$ to suppress the severely degraded regions in $\bm{F}_{vi}$ and replace them with the corresponding information from $\hat{\bm{F}}_{ir}$. The specific process can be formulated as
\begin{equation}
	\hat{\bm{F}}_{vi} = TF\!\left( \hat{\bm{F}}_{ir} \odot \bm{H} + \bm{F}_{vi} \odot \left( \mathbf{1} - \bm{H} \right) + \bm{F}_{vi} \right),
\end{equation}
where $TF$ denotes the Transformer block. To ensure the quality of $\hat{\bm{F}}_{vi}$, it is passed through a $3 \times 3$ convolution to obtain the dehazed image $\hat{\bm{I}}_{vi}$, and an $L_1$ loss is used to optimize the network:
\begin{equation}
	\ell_1 = \left\| \hat{\bm{I}}_{vi} - \bm{I}_{vi,gt} \right\|_1,
\end{equation}
where $\bm{I}_{vi,gt}$ represents the corresponding ground-truth haze-free visible image.

\subsection{Multi-stage Prompt Embedding Fusion}
With the assistance of infrared features ${\bm{\hat{F}}}_{ir}$, we obtain the dehazed visible image features ${\bm{\hat{F}}}_{vi}$. These features are then fused with the infrared image features ${\bm{F}}_{ir}$ to reconstruct a dehazed fusion result. This approach enables us to fuse hazy IR-VIS images within a single framework, producing a dehazed fusion output. In this process, an effective fusion method is essential to achieve high-quality fusion results. To prevent residual haze in ${\bm{\hat{F}}}_{vi}$ from affecting the fusion, we propose the MsPE-FM.

As shown in Fig.~\ref{Fig2}, in the MsPE-FM, the restored feature $\bm{\hat{F}}_{vi}$ and the initial feature $\bm{F}_{ir}$ are input into the PGM to obtain the prompt embedding $\bm{\hat P}_{vi}^1$ at the first stage of the fusion process. After concatenating $\bm{\hat{F}}_{vi}$ and $\bm{F}_{ir}$, the result is passed through the Restormer \cite{33}. This output, along with $\bm{\hat P}_{vi}^1$, is then input into the PEB to obtain the fusion result for the next stage. In the second stage, $\bm{\hat P}_{vi}^1$ is first passed through a $1 \times 1$ Conv layer to adjust the number of channels, resulting in the adjusted prompt embedding $\bm{\hat P}_{vi}^2$, which adapts to the changes in feature channels during the second-stage feature extraction. In this fusion process, five fusion blocks, each consisting of prompt embeddings, a Restormer, and a PEB, are used to achieve the fusion of IR-VIS features. Within these fusion blocks, two residual connections are employed to prevent information loss. Finally, the fused features pass through a $1 \times 1$ Conv layer to reconstruct the fused result $\bm{I}_f$.

To ensure that the gradients of the fusion result are consistent with those of the input infrared image and the clear visible image across the three RGB channels, we employ the gradient loss from \cite{35} to optimize the parameters of the entire network:
\begin{equation}
	{\ell_\nabla} = \frac{1}{{HW}} \sum_{i=1}^3 \left\| \nabla {\bm{I}}_{f}^{i} - \max \left( \left| \nabla {\bm{I}}_{ir}\right|, \left| \nabla {\bm{I}}_{vi,gt}^{i}\right| \right) \right\|_1
\end{equation}
where $\nabla$ denotes the gradient operator, and $i$ represents the R, G, B channels. Additionally, to ensure that the fused image maintains consistent pixel intensity with both the IR and VIS images, we utilize a pixel intensity consistency loss function $\ell_{int}$ to update the network parameters: 
\begin{equation}
	{\ell_{int}} = \frac{1}{{HW}} \sum_{i=1}^3 \left\| {\bm{I}}_{f}^{i} - \max \left( {\bm{I}}_{ir}, {\bm{I}}_{vi,gt}^{i}\right) \right\|_1
\end{equation}
The total loss is then formulated as:
\begin{equation}
	{\ell_{total}} = {\ell_{int}} + {\ell_\nabla} + \alpha {\ell_1}
\end{equation}
where $\alpha$ is a hyperparameter that adjusts the contribution of the $L_1$-loss in this optimization process.

\section{Experiments}\label{section:experiments}
\subsection{Experimental Configurations}
\textbf{Dataset}. In this work, we utilize 1,083 IR-VIS image pairs from the MSRS dataset \cite{36} as the training set. This dataset includes a wide variety of scenes, such as vehicles, pedestrians, houses, and streets, offering a rich and diverse set of visual data for training purposes. For testing, we use 361 image pairs from the MSRS dataset for both qualitative and quantitative comparative experiments, ensuring no overlap with the training set. Additionally, we assess the effectiveness and generalization capability of our method on 100 image pairs from the M$^3$FD dataset \cite{16} and 50 image pairs from the RoadScene dataset \cite{38}. The M$^3$FD dataset contains IR-VIS image pairs from scenes such as university campuses, vacation spots, and urban main roads, while the RoadScene dataset includes IR-VIS image pairs selected from the representative scenes including pedestrians, vehicles, roads, and buildings. To generate hazy image pairs, we apply the atmospheric scattering model \cite{34} to introduce haze into the visible images in both the training and test sets.

\textbf{Metrics}. To objectively evaluate the fusion performance of different methods, we adopt five commonly used image quality assessment metrics: Mutual Information ($Q_{MI}$) \cite{40}, Gradient-based Fusion Performance ($Q_{AB/F}$) \cite{41}, Chen-Varshney Metric ($Q_{CV}$) \cite{42}, Sum of Correlation of Differences ($Q_{SCD}$) \cite{43}, and Visual Information Fidelity ($Q_{VIF}$) \cite{44}. These metrics are used to assess the quality of the fusion results, with clear source images (without haze) as reference images when necessary. Additionally, to evaluate the perceptual quality of the dehazing effects within the fusion results, we employ the Perceptual Index ($Q_{PI}$) \cite{45}, Natural Image Quality Evaluator ($Q_{NIQE}$) \cite{46}, and Spatial Frequency ($Q_{SF}$) \cite{47}. Specifically, $Q_{PI}$ and $Q_{NIQE}$ are widely used no-reference naturalness-based perceptual metrics that reflect the overall visual quality and naturalness of restored images, while $Q_{SF}$ measures spatial frequency and thus reflects edge sharpness and the richness of spatial details. According to the evaluation criteria, lower $Q_{CV}$, $Q_{PI}$, and $Q_{NIQE}$ values indicate better fusion performance, while higher values for the remaining metrics signify improved quality.

\subsection{Implementation Details}

All experiments are conducted using the PyTorch framework on a single 24GB NVIDIA GeForce RTX 4090 GPU. In our implementation, the entire network is trained in an end-to-end manner under the joint restoration and fusion loss, so that both branches can be optimized collaboratively. During training, images are randomly cropped to $256 \times 256$ patches, with data augmentation techniques such as horizontal and vertical flipping applied. The model is trained for a total of 300 epochs, using a batch size of 6 and the AdamW optimizer \cite{39}. The initial learning rate is set to $2 \times 10^{-4}$ and is gradually reduced to $2 \times 10^{-6}$ following a cosine annealing schedule.

\subsection{Comparison with State-of-the-art Methods}
In order to verify the effectiveness of our method, we compare it with two existing methods. The first methodology involves initially applying advanced image dehazing algorithms to remove haze from the visible images, followed by fusing the dehazed images with the infrared images. For this purpose, we select the latest and most effective dehazing methods, namely DIACMP \cite{48} and Dehazeformer \cite{49}. Next, we apply representative IR-VIS image fusion methods, such as MLFusion \cite{28}, U2Fusion \cite{38}, LRRNet \cite{5}, ALFusion \cite{3}, TIMFusion \cite{50}, MRFS \cite{51}, SHIP  \cite{ship} and FreeFusion \cite{freefusion}, to fuse the dehazed visible images with the infrared images. The second methodology employs the Text-IF method \cite{31}, which directly restores and fuses hazy images with the assistance of text information.

\begin{figure*}[t!]
	\centering
	\includegraphics[width=1.0\textwidth]{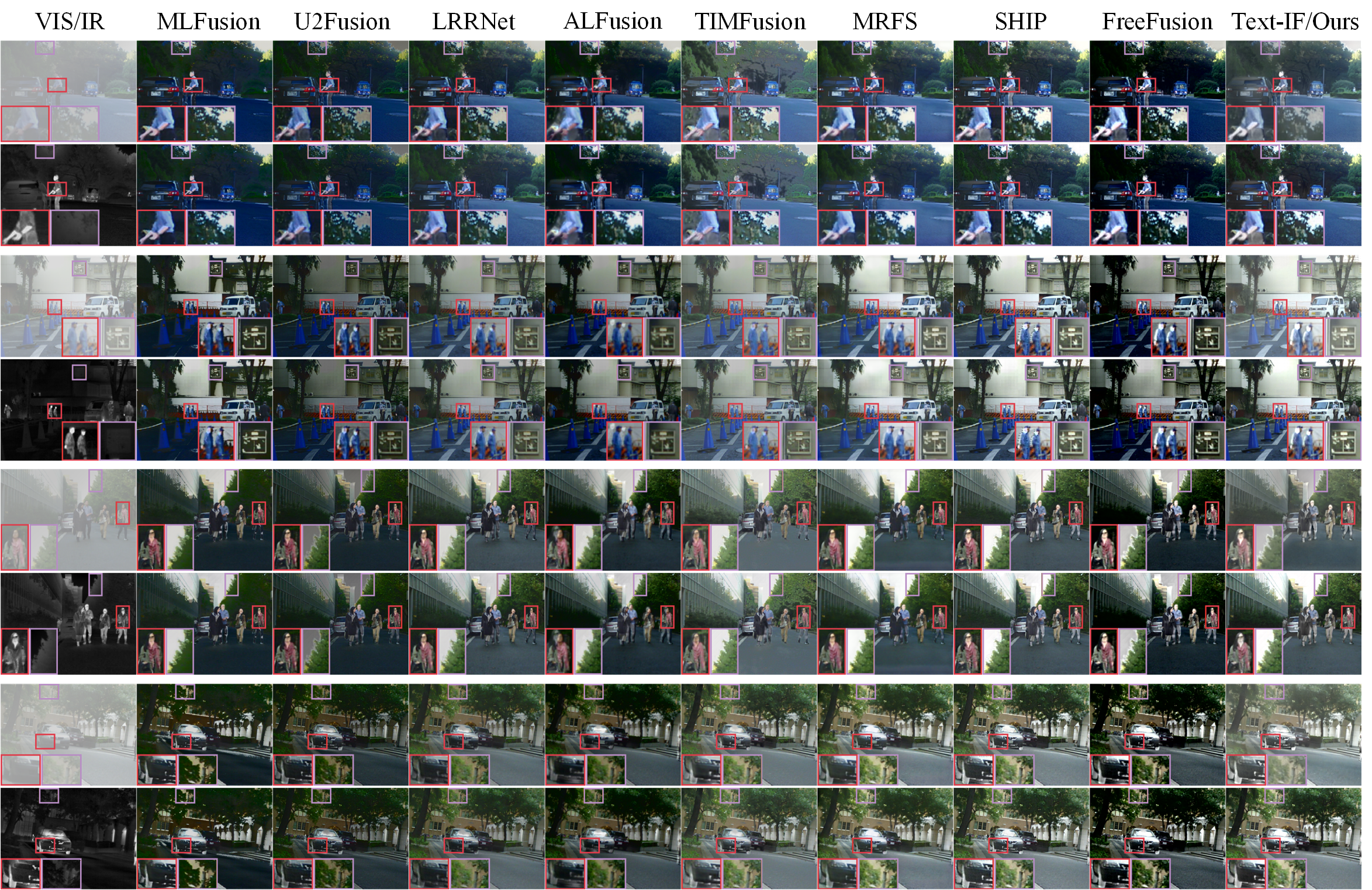}\vspace{-2mm}
	\caption{Visual comparison of fusion results from different methods on the MSRS dataset. In the fusion results generated by the comparison methods, except for the last column, the first row of each image pair represents the result of first dehazing with DIACMP and then fusing. The second row shows the result of dehazing with Dehazeformer and then fusing. The last column represents the result of fusion using Text-IF (first row) and our method (second row).
	}\vspace{-2mm}
	\label{fig3}
\end{figure*}

\textbf{Experiments on MSRS dataset.}
To intuitively evaluate the fusion performance of different algorithms on the MSRS dataset, four pairs of IR-VIS images are selected, as shown in Fig. \ref{fig3}. As indicated from the red boxes in the first and second rows of Fig. \ref{fig3}, our method effectively preserves thermal radiation information, clearly highlighting the trousers of pedestrians, which most other methods fail to achieve. Although TIMFusion and Text-IF can also accomplish this to some extent, the results within the purple boxes reveal that they fail to accurately restore the texture details of trees, resulting in blurred outputs. In the red boxes of the third and fourth rows, our method preserves the details of windows while maintaining the scene brightness, producing a clear and well-restored window. In contrast, MLFusion, U2Fusion, and ALFusion suffer from brightness loss, leading to blurred scenes and poor visual effects. LRRNet, TIMFusion, MRFS, and Text-IF fail to deliver satisfactory contrast.  Moreover, the other two sets of experimental results shown in Fig. \ref{fig3} further highlight that the proposed method achieves superior visual performance compared to the competing methods.

Additionally, we conduct a quantitative comparison on 361 image pairs from the MSRS dataset to verify the effectiveness of our method. Tables \ref{tab1} and \ref{tab2} present the experimental results based on the DIACMP and Dehazeformer dehazing methods, respectively. As shown in Tables \ref{tab1}--\ref{tab2}, our method ranks first across all eight metrics, demonstrating its outstanding performance on the MSRS dataset. The higher $Q_{AB/F}$ and $Q_{CV}$ scores indicate that our fused images achieve superior detail clarity. Meanwhile, the best $Q_{SCD}$ and $Q_{VIF}$ scores suggest that our method ensures better visual consistency, closely matching the clear source images. The $Q_{MI}$ metric reflects the shared information between the fused and source images, confirming that our approach preserves more source image information. As perceptual quality metrics, $Q_{NIQE}$ and $Q_{PI}$ evaluate image naturalness and perceptual quality in a no-reference manner, where lower values indicate better alignment with human visual perception. Furthermore, $Q_{SF}$ measures the sharpness of image edges, demonstrating that our method produces haze-free fused images that are both natural and sharp.

\newcommand{\best}[1]{\colorbox{red!15}{#1}}
\newcommand{\second}[1]{\colorbox{blue!15}{#1}}
\begin{table}[ht!]
	\centering
	\caption{Quantitative analysis of our method compared with DIACMP+fusion and text-if on the MSRS dataset. The best and second-best performances are highlighted with \best{Red} and \second{Blue} backgrounds, respectively.}
	\renewcommand\arraystretch{1.2}
	{\footnotesize\centerline{\tabcolsep=8pt
			\begin{tabular}{ccccccccccc}
				\hline
				\hline 
				\textbf{Methods} & $Q_{MI}$$\uparrow$ & $Q_{AB/F}$$\uparrow$ & $Q_{CV}$$\downarrow$ & $Q_{SCD}$$\uparrow$ & $Q_{VIF}$$\uparrow$ & $Q_{NIQE}$$\downarrow$ & $Q_{PI}$$\downarrow$ & $Q_{SF}$$\uparrow$ \\ 
				\hline
				MLFusion & 1.706 & 0.274 & 608.391 & 1.121 & 0.199 & 5.022 & 4.285 & \second{9.953} \\ 
				U2Fusion & 1.301 & 0.330 & 838.330 & \second{1.364} & 0.226 & 4.977 & 4.354 & 7.209 \\ 
				LRRNet & 1.864 & 0.441 & 614.101 & 1.000 & 0.288 & 4.750 & 4.333 & 8.479 \\ 
				ALFusion & 1.449 & 0.377 & 700.180 & 1.201 & 0.257 & 5.391 & 5.419 & 7.417 \\ 
				TIMFusion & 1.800 & 0.382 & 1048.770 & 1.090 & 0.295 & 4.515 & \second{4.256} & 8.136 \\ 
				MRFS & 1.580 & 0.457 & \second{332.068} & 1.348 & 0.325 & 5.068 & 4.800 & 9.042 \\ 
				\textcolor{blue}{SHIP} & \second{2.021} & 0.491 & 509.112 & 1.316 & 0.338 & 5.287 & 4.723 & 8.512 \\
				\textcolor{blue}{FreeFusion} & 1.892 & 0.512 & 441.721 & 1.214 & 0.287 & 4.996 & 4.478 & 7.783 \\
				Text-IF & 1.428 & \second{0.558} & 458.252 & 1.351 & \second{0.369} & \second{4.260} & 4.312 & 8.393 \\ 
				Ours & \best{2.720} & \best{0.652} & \best{238.420} & \best{1.662} & \best{0.490} & \best{4.110} & \best{3.811} & \best{11.050} \\ 
				\hline
	\end{tabular}}}
	\label{tab1}
\end{table}

\begin{table}[h!]
	\centering
	\caption{Quantitative analysis of our method compared with dehazeformer+fusion and text-if on the MSRS dataset. The best and second-best performances are highlighted with \best{Red} and \second{Blue} backgrounds, respectively.}
	\renewcommand\arraystretch{1.2}
	{\footnotesize\centerline{\tabcolsep=8pt
			\begin{tabular}{ccccccccccc}
				\hline
				\hline 
				\textbf{Methods} & $Q_{MI}$$\uparrow$ & $Q_{AB/F}$$\uparrow$ & $Q_{CV}$$\downarrow$ & $Q_{SCD}$$\uparrow$ & $Q_{VIF}$$\uparrow$ & $Q_{NIQE}$$\downarrow$ & $Q_{PI}$$\downarrow$ & $Q_{SF}$$\uparrow$ \\ 
				\hline
				MLFusion & 1.720 & 0.270 & 613.291 & 1.119 & 0.199 & 5.052 & 4.337 & \second{9.826} \\ 
				U2Fusion & 1.311 & 0.324 & 832.878 & \second{1.361} & 0.225 & 5.185 & 4.439 & 7.071 \\ 
				LRRNet & 1.883 & 0.436 & 608.019 & 0.999 & 0.288 & 4.883 & 4.445 & 8.317 \\ 
				ALFusion & 1.471 & 0.370 & 695.816 & 1.198 & 0.256 & 5.468 & 5.542 & 7.339 \\ 
				TIMFusion & 1.785 & 0.382 & 1057.931 & 1.092 & 0.299 & 4.567 & 4.355 & 8.303 \\ 
				MRFS & 1.603 & 0.448 & \second{328.901} & 1.340 & 0.323 & 5.150 & 4.905 & 8.843 \\
				\textcolor{blue}{SHIP} & \second{2.133} & 0.486 & 504.067 & 1.249 & 0.328 & 5.266 & 4.679 & 8.448 \\
				\textcolor{blue}{FreeFusion} & 1.965 & 0.498 & 438.963 & 1.196 & 0.276 & 4.985 & 4.556 & 7.675 \\ 
				Text-IF & 1.428 & \second{0.558} & 458.252 & 1.351 & \second{0.369} & \second{4.260} & \second{4.312} & 8.393 \\ 
				Ours & \best{2.720} & \best{0.652} & \best{238.420} & \best{1.662} & \best{0.490} & \best{4.110} & \best{3.811} & \best{11.050} \\ 
				\hline
	\end{tabular}}}
	\label{tab2}
\end{table}

\textbf{Experiments on M$^3$FD dataset.}
\begin{figure*}[t!]
	\centering
	\includegraphics[width=1.0\textwidth]{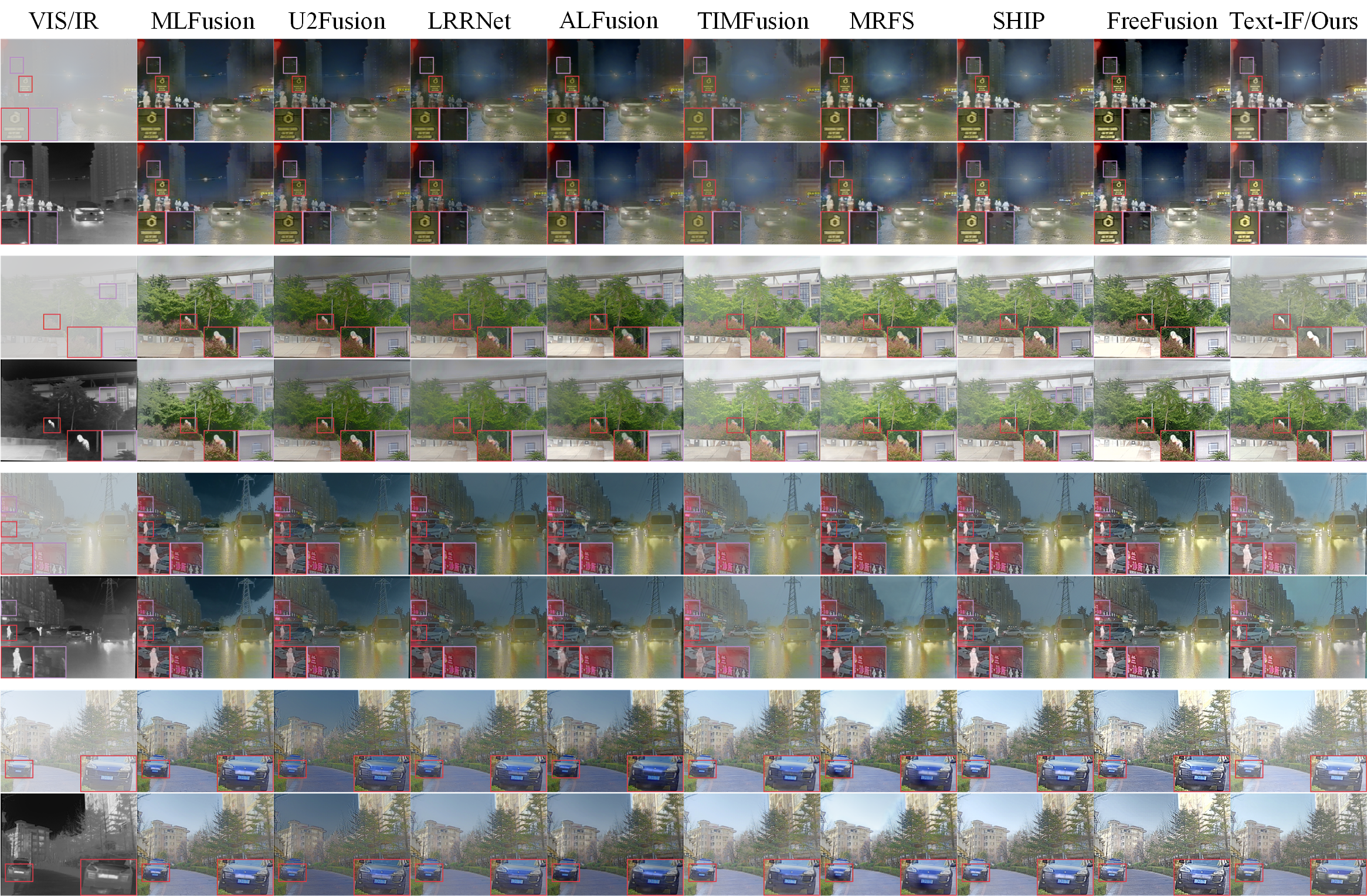}\vspace{-2mm}
	\caption{Visual comparison of fusion results from different methods on the M$^3$FD dataset. In the fusion results generated by the comparison methods, except for the last column, the first row of each image pair represents the result of first dehazing with DIACMP and then fusing. The second row shows the result of dehazing with Dehazeformer and then fusing. The last column represents the result of fusion using Text-IF (first row) and our method (second row).}
	\vspace{-2mm}
	\label{fig4}
\end{figure*}
To evaluate the generalization ability of our method on the M$^3$FD dataset, we select four pairs of IR-VIS images, with the visualization results shown in the Fig. \ref{fig4}. As illustrated in the first and second rows of Fig. \ref{fig4}, most methods exhibit blurring on the store signs, whereas our method preserves clear edges and texture details. Although MLFusion is able to preserve the texture information of the signs to some extent, it performs poorly in restoring background details, resulting in noticeable blurring in the background regions. This issue can be clearly observed in the building areas on both the left and right sides of the first image pair. In these regions, the results produced by MLFusion appear overly smooth, with reduced contrast in wall surfaces and window-frame details. This issue is also evident in the second and third image pairs, where MLFusion and other methods display color distortion in the sky regions. Thanks to the incorporation of infrared information during the dehazing stage in our method, the nextwork effectively restores these sky regions, achieving superior restoration and fusion results.
\begin{table}[ht!]
	\centering
	\caption{Quantitative analysis of our method compared with diacmp+fusion and text-if on the M$^3$FD dataset. The best and second-best performances are highlighted with \best{Red} and \second{Blue} backgrounds, respectively.}
	\renewcommand\arraystretch{1.2}
	{\footnotesize\centerline{\tabcolsep=8pt
			\begin{tabular}{ccccccccc}
				\hline
				\hline 
				\textbf{Methods} & $Q_{MI}$$\uparrow$ & $Q_{AB/F}$$\uparrow$ & $Q_{CV}$$\downarrow$ & $Q_{SCD}$$\uparrow$ & $Q_{VIF}$$\uparrow$ & $Q_{NIQE}$$\downarrow$ & $Q_{PI}$$\downarrow$ & $Q_{SF}$$\uparrow$ \\ 
				\hline
				MLFusion & 1.842 & 0.411 & 937.069 & 1.134 & 0.390 & 4.840 & 3.230 & 12.025 \\ 
				U2Fusion & 1.607 & 0.511 & 805.163 & \second{1.266} & 0.358 & 4.908 & 3.377 & 12.009 \\ 
				LRRNet & 1.579 & 0.473 & 724.325 & 1.195 & 0.358 & 4.355 & 3.245 & 11.529 \\ 
				ALFusion & 1.510 & 0.419 & 890.521 & 1.205 & 0.322 & 4.878 & 3.960 & 9.291 \\ 
				TIMFusion & 1.772 & 0.445 & 859.554 & 0.921 & 0.340 & \second{4.249} & \second{3.228} & 11.023 \\ 
				MRFS & 1.756 & 0.514 & 666.466 & 1.208 & 0.415 & 4.304 & 3.277 & 13.071 \\  
				\textcolor{blue}{SHIP} & 1.929 & 0.529 & 676.327 & 1.258 & 0.424 & 4.657 & 3.475 & 11.920 \\
				\textcolor{blue}{FreeFusion} & 1.865 & 0.538 & 986.925 & 1.272 & 0.361 & 4.437 & 3.235 & \second{13.760} \\ 
				Text-IF & \second{1.935} & \second{0.542} & \best{634.406} & 1.189 & \second{0.427} & 5.277 & 3.902 & 11.940 \\ 
				Ours & \best{2.070} & \best{0.571} & \second{640.661} & \best{1.357} & \best{0.440} & \best{4.144} & \best{3.158} & \best{14.140} \\ 
				\hline
	\end{tabular}}}
	\label{tab3}
\end{table}

\begin{table}[htbp]
	\centering
	\caption{Quantitative analysis of our method compared with dehazeformer+fusion and text-if on the M$^3$FD dataset. The best and second-best performances are highlighted with \best{Red} and \second{Blue} backgrounds, respectively.}
	\renewcommand\arraystretch{1.2}
	{\footnotesize\centerline{\tabcolsep=8pt
			\begin{tabular}{ccccccccccc}
				\hline
				\hline 
				\textbf{Methods} & $Q_{MI}$$\uparrow$ & $Q_{AB/F}$$\uparrow$ & $Q_{CV}$$\downarrow$ & $Q_{SCD}$$\uparrow$ & $Q_{VIF}$$\uparrow$ & $Q_{NIQE}$$\downarrow$ & $Q_{PI}$$\downarrow$ & $Q_{SF}$$\uparrow$ \\ 
				\hline
				MLFusion & 1.879 & 0.400 & 1023.684 & 1.133 & 0.393 & 4.639 & \second{3.232} & 11.510 \\ 
				U2Fusion & 1.647 & 0.497 & 824.810 & \second{1.281} & 0.355 & 5.251 & 3.635 & 11.295 \\ 
				LRRNet & 1.592 & 0.463 & 737.595 & 1.206 & 0.360 & 4.528 & 3.455 & 11.116 \\ 
				ALFusion & 1.503 & 0.403 & 922.165 & 1.227 & 0.324 & 5.095 & 4.196 & 8.974 \\ 
				TIMFusion & 1.719 & 0.449 & 814.112 & 0.919 & 0.336 & \second{4.420} & 3.357 & 11.325 \\ 
				MRFS & 1.788 & 0.509 & 693.528 & 1.200 & 0.417 & 4.467 & 3.460 & 12.551 \\
				\textcolor{blue}{SHIP} & 1.908 & 0.525 & 687.446 & 1.239 & 0.423 & 4.779 & 3.578 & 11.720 \\
				\textcolor{blue}{FreeFusion} & 1.860 & 0.533 & 990.661 & 1.233 & 0.358 & 4.530 & 3.339 & \second{13.640} \\  
				Text-IF & \second{1.935} & \second{0.542} & \best{634.406} & 1.189 & \second{0.427} & 5.277 & 3.902 & 11.940 \\ 
				Ours & \best{2.070} & \best{0.571} & \second{640.661} & \best{1.357} & \best{0.440} & \best{4.144} & \best{3.158} & \best{14.140} \\ 
				\hline
	\end{tabular}}}
	\label{tab4}
\end{table}

In terms of preserving thermal target information, our method also demonstrates a leading performance, which is particularly evident in the results shown in the fifth and sixth rows. Methods such as U2Fusion, LRRNet, ALFusion, and TIMFusion fail to retain thermal target information, leading to fusion results that do not effectively highlight thermal targets. In contrast, our method adopts a multi-stage prompt information injection strategy during the fusion phase, ensuring the infrared information is well-preserved. Although MLFusion, MRFS, and Text-IF can also emphasize targets to some extent, their performance in restoring background details remains suboptimal.

Quantitative comparison results on M$^3$FD test set, based on the DIACMP and Dehazeformer dehazing methods, are presented in Tables \ref{tab3} and \ref{tab4}, respectively. It can be observed that our proposed method ranks first in seven evaluation metrics and second in $Q_{CV}$, indicating its superior restoration and fusion capabilities.

\textbf{Experiments on RoadScene dataset.}
\begin{figure*}[t!]
	\centering
	\includegraphics[width=1.0\textwidth]{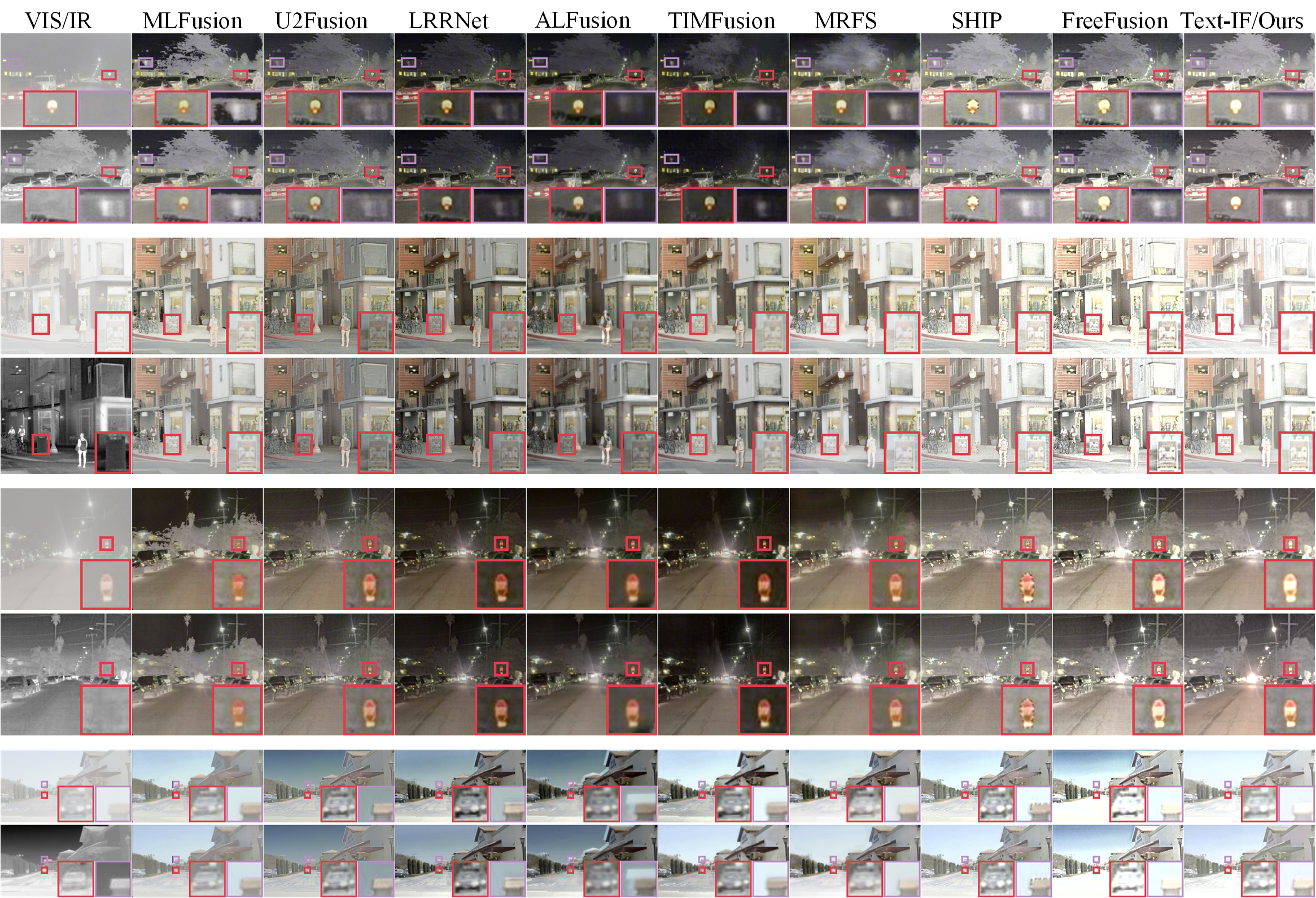}\vspace{-2mm}
	\caption{Visual comparison of fusion results from different methods on the RoadScene dataset. In the fusion results generated by the comparison methods, except for the last column, the first row of each image pair represents the result of first dehazing with DIACMP and then fusing. The second row shows the result of dehazing with Dehazeformer and then fusing. The last column represents the result of fusion using Text-IF (first row) and our method (second row).}\vspace{-2mm}
	\label{fig5}
\end{figure*}
We select four pairs of IR-VIS images from the RoadScene dataset to further evaluate the effectiveness and generalization capability of our method. The qualitative comparison results are presented in the Fig. \ref{fig5}.  As shown in the first set of results in Fig. \ref{fig5}, our approach provides sharper object edges and more detailed textures. In the third and fourth rows, the content on the billboard within the red box is significantly clearer in our method compared to others, where varying degrees of blurriness are observed. Notably, the content displayed by Text-IF is completely unrecognizable.The results in the fifth and sixth rows indicate that our method can generate fused images with high contrast, preserving the original colors of the signboards while maintaining clear edges.  From the results in the red box in the seventh and eighth rows, it can be seen that our method clearly highlights the information of distant vehicles, whereas the results generated by MLFusion, MRFS, and Text-IF are relatively blurry. Although other comparison methods can somewhat enhance vehicle information, as shown in the results in the purple box, their ability to restore the distant sky and roof areas is inferior to that of our method.
\begin{table}[ht!]
	\centering
	\caption{Quantitative analysis of our method compared with diacmp+fusion and text-if on the RoadScene dataset. The best and second-best performances are highlighted with \best{Red} and \second{Blue} backgrounds, respectively.}
	\renewcommand\arraystretch{1.2}
	{\footnotesize\centerline{\tabcolsep=8pt
			\begin{tabular}{ccccccccc}
				\hline
				\hline
				\textbf{Methods} & $Q_{MI}$$\uparrow$ & $Q_{AB/F}$$\uparrow$ & $Q_{CV}$$\downarrow$ & $Q_{SCD}$$\uparrow$ & $Q_{VIF}$$\uparrow$ & $Q_{NIQE}$$\downarrow$ & $Q_{PI}$$\downarrow$ & $Q_{SF}$$\uparrow$ \\ 
				\hline
				MLFusion & \second{2.350} & 0.457 & 509.388 & 1.248 & 0.411 & 3.703 & 3.196 & 12.083 \\ 
				U2Fusion & 1.852 & 0.492 & 842.539 & 1.201 & 0.342 & 3.856 & \second{2.960} & 12.650 \\ 
				LRRNet & 1.947 & 0.363 & 622.880 & 0.827 & 0.359 & \second{3.669} & 3.009 & 12.498 \\ 
				ALFusion & 1.753 & 0.327 & 831.380 & 0.956 & 0.273 & 4.311 & 4.191 & 8.793 \\ 
				TIMFusion & 2.349 & 0.356 & 699.810 & 0.881 & 0.425 & 4.484 & 4.241 & 10.307 \\ 
				MRFS & 2.115 & 0.391 & \second{477.748} & 1.362 & 0.393 & 3.932 & 3.615 & 11.269 \\
				\textcolor{blue}{SHIP} & 2.309 & 0.495 & 549.782 & 1.298 & \second{0.435} & 3.695 & 3.644 & 13.346 \\
				\textcolor{blue}{FreeFusion} & 2.278 & 0.449 & 500.116 & 1.294 & 0.401 & 4.552 & 3.278 & 12.847 \\ 
				Text-IF & 2.256 & \best{0.583} & 497.001 & \best{1.456} & 0.418 & 3.737 & 3.049 & \second{13.746} \\ 
				Ours & \best{2.673} & \second{0.511} & \best{454.497} & \second{1.399} & \best{0.446} & \best{3.333} & \best{2.806} & \best{14.804} \\ 
				\hline
	\end{tabular}}}
	\label{tab5}
\end{table}

\begin{table}[ht!] 
	\centering
	\caption{Quantitative analysis of our method compared with dehazeformer+fusion and text-if on the RoadScene dataset. The best and second-best performances are highlighted with \best{Red} and \second{Blue} backgrounds, respectively.}
	\renewcommand\arraystretch{1.2}
	{\footnotesize\centerline{\tabcolsep=8pt
			\begin{tabular}{ccccccccccc}
				\hline
				\hline 
				\textbf{Methods} & $Q_{MI}$$\uparrow$ & $Q_{AB/F}$$\uparrow$ & $Q_{CV}$$\downarrow$ & $Q_{SCD}$$\uparrow$ & $Q_{VIF}$$\uparrow$ & $Q_{NIQE}$$\downarrow$ & $Q_{PI}$$\downarrow$ & $Q_{SF}$$\uparrow$ \\ 
				\hline
				MLFusion & \second{2.425} & 0.463 & 509.536 & 1.239 & 0.411 & 3.795 & 3.343 & 11.364 \\ 
				U2Fusion & 1.864 & 0.495 & 862.820 & 1.175 & 0.340 & 3.873 & 3.029 & 12.117 \\ 
				LRRNet & 1.948 & 0.360 & 617.051 & 0.780 & 0.357 & \second{3.534} & \second{2.967} & 12.326 \\ 
				ALFusion & 1.726 & 0.315 & 844.762 & 0.934 & 0.257 & 4.321 & 4.258 & 8.545 \\ 
				TIMFusion & 2.359 & 0.354 & 694.572 & 0.869 & 0.415 & 4.457 & 4.268 & 10.099 \\ 
				MRFS & 2.119 & 0.396 & \second{472.542} & 1.332 & 0.389 & 3.847 & 3.619 & 10.922 \\
				\textcolor{blue}{SHIP} & 2.237 & 0.461 & 539.668 & 1.286 & \second{0.426} & 3.455 & 3.646 & 13.238 \\
				\textcolor{blue}{FreeFusion} & 2.168 & 0.427 & 484.662 & 1.281 & 0.392 & 4.458 & 3.169 & 12.647 \\  
				Text-IF & 2.256 & \best{0.583} & 497.001 & \best{1.456} & 0.418 & 3.737 & 3.049 & \second{13.746} \\ 
				Ours & \best{2.673} & \second{0.511} & \best{454.497} & \second{1.399} & \best{0.446} & \best{3.333} & \best{2.806} & \best{14.804} \\ 
				\hline
	\end{tabular}}}
	\label{tab6}
\end{table}

Quantitative comparison results on the RoadScene dataset are presented in Tables \ref{tab5} and \ref{tab6}, which utilize the DIACMP and Dehazeformer dehazing methods respectively. Our method ranks first in six evaluation metrics, with $Q_{AB/F}$  and  $Q_{SCD}$ ranking second. The best $Q_{NIQE}$  and $Q_{PI}$ scores indicate that our method generates fused images that are both natural and sharp.

\begin{figure*}[ht!]
	\centering
	\includegraphics[width=1.0\textwidth]{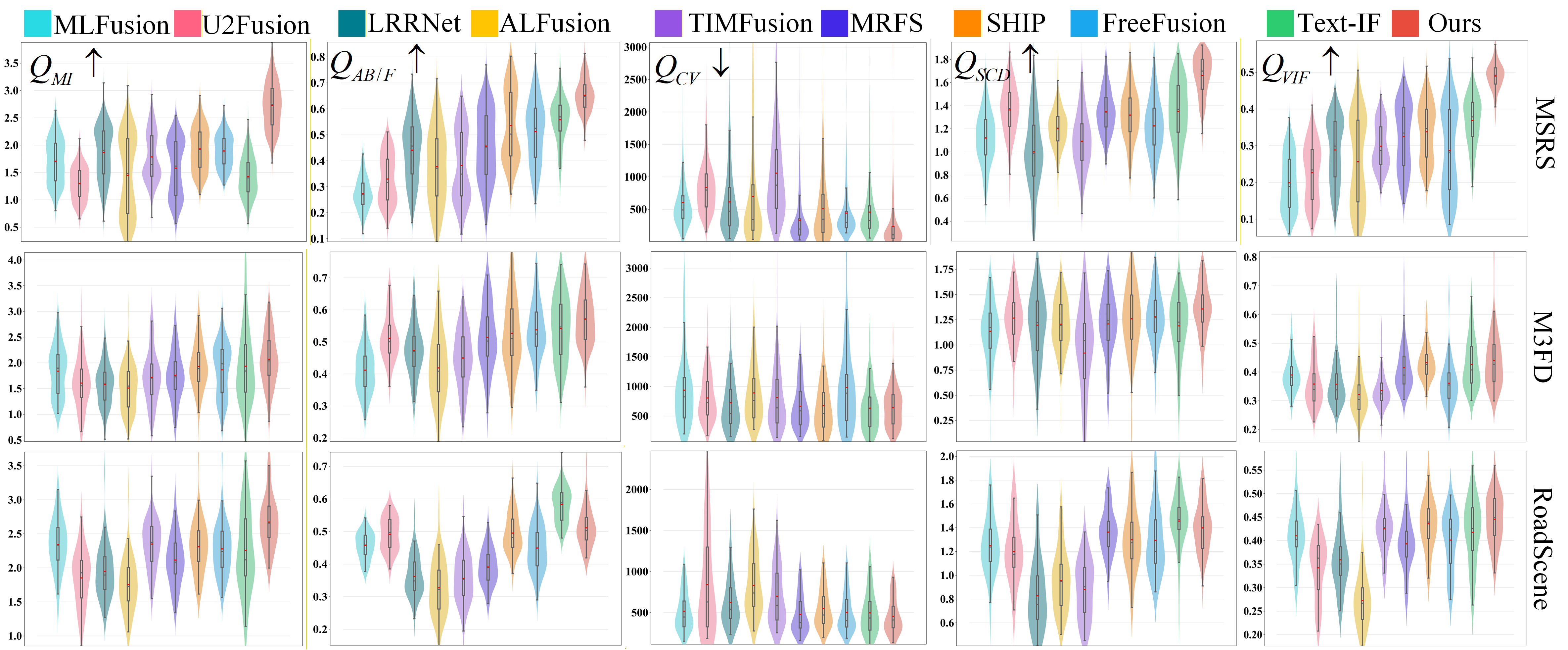}\vspace{-2mm}
	\caption{Visualization of objective evaluation results.  the group shows the results of using DIACMP for dehazing followed by fusion on MSRS, M$^3$FD, and RoadScene.}\vspace{-2mm}
	\label{fig6}
\end{figure*}

\begin{figure*}[ht!]
	\centering
	\includegraphics[width=1.0\textwidth]{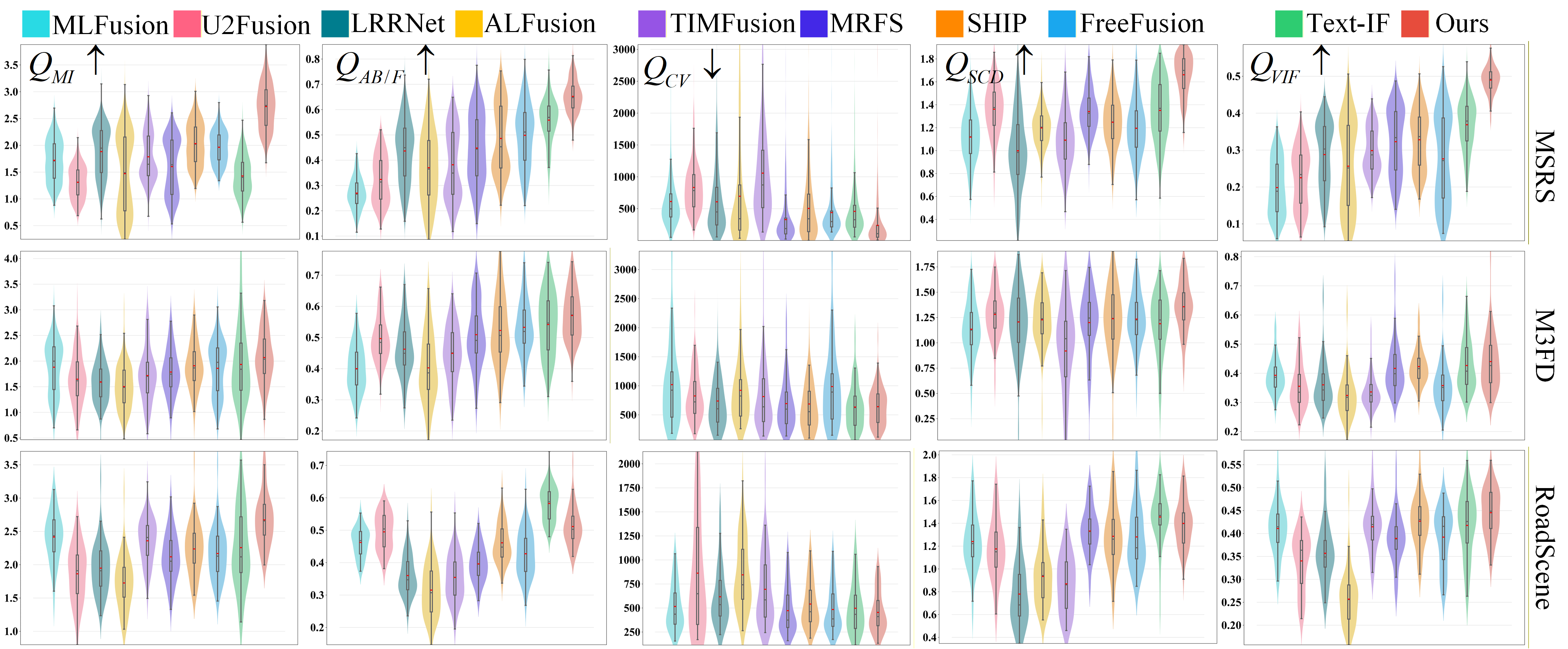}\vspace{-2mm}
	\caption{Visualization of objective evaluation results.  the group shows the results of using Dehazeformer for dehazing followed by fusion on MSRS, M$^3$FD, and RoadScene.}\vspace{-2mm}
	\label{fig7}
\end{figure*}

\begin{figure*}[ht!]
	\centering
	\includegraphics[width=0.8\textwidth]{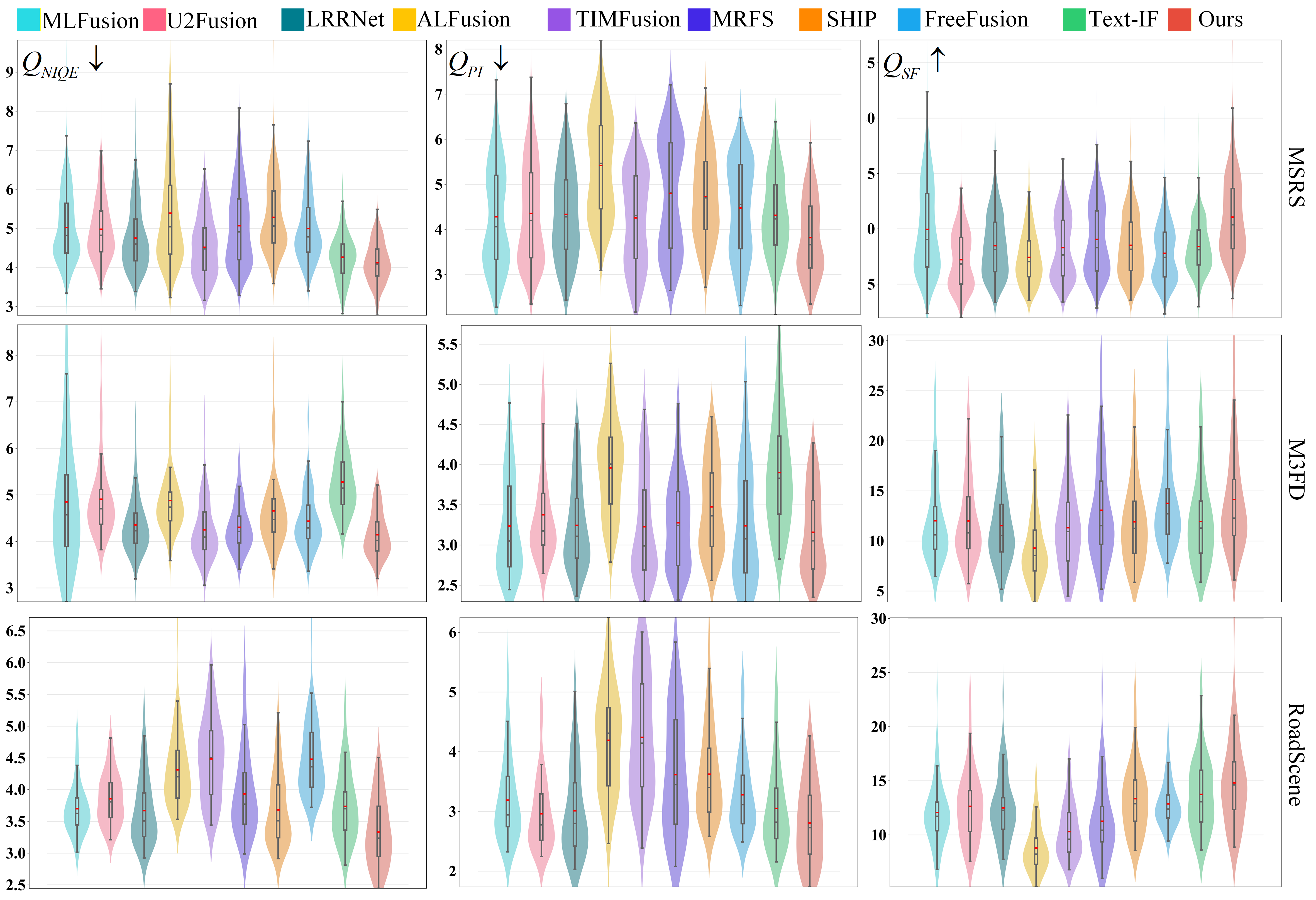}
	\caption{Visualization of objective evaluation results. the group presents the objective evaluation results of fusion images obtained using DIACMP dehazing method, assessed with haze evaluation metrics.}\vspace{-2mm}
	\label{fig8}
\end{figure*}

\begin{figure*}[ht!]
	\centering
	\includegraphics[width=0.8\textwidth]{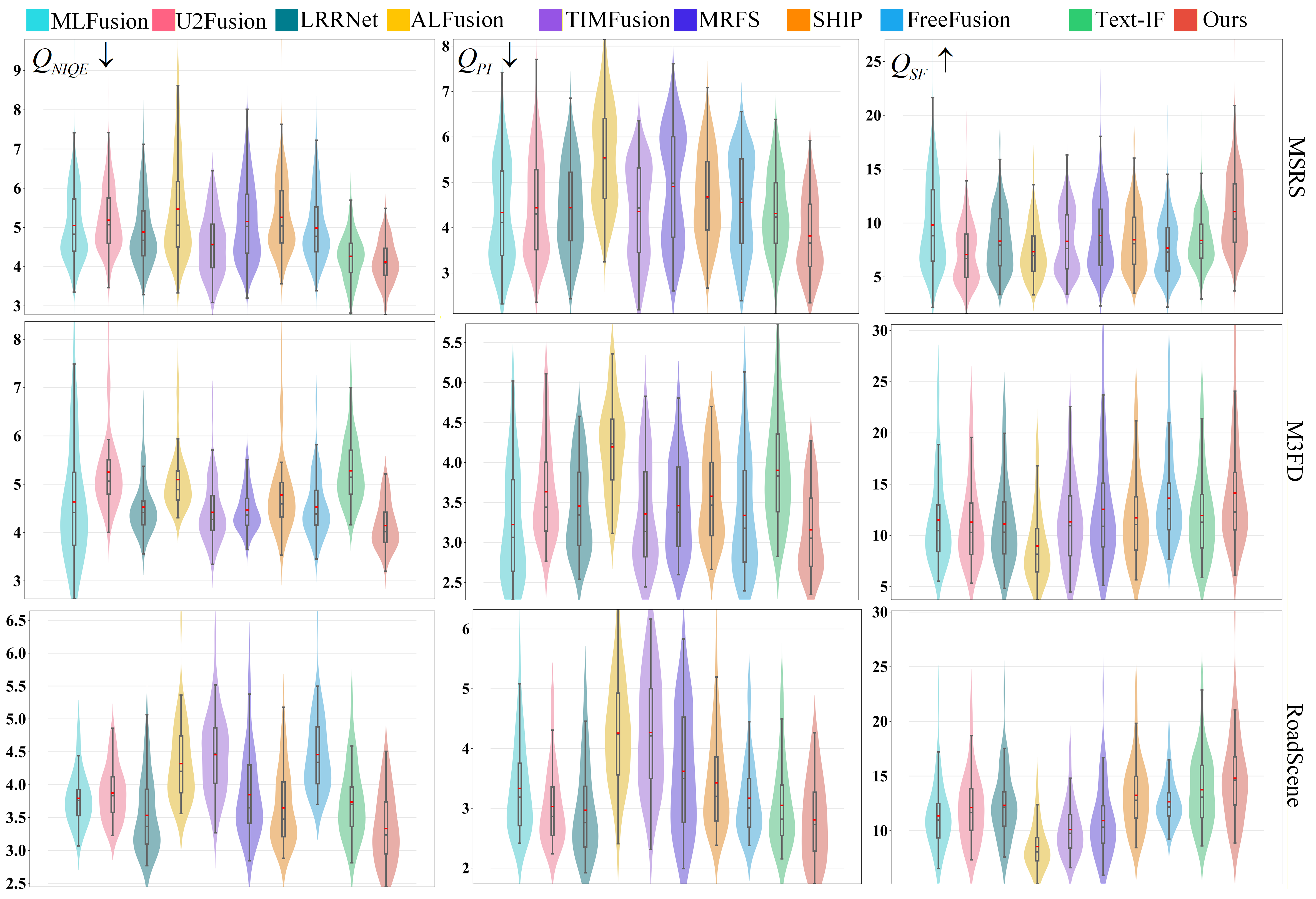}
	\caption{Visualization of objective evaluation results.   the group presents the objective evaluation results of fusion images obtained using Dehazeformer dehazing method, assessed with haze evaluation metrics.}\vspace{-2mm}
	\label{fig9}
\end{figure*}

To further validate the advantages of our method, we present violin plots for quantitative comparison in Figs. \ref{fig6}–\ref{fig9}. These plots combine data density distribution (represented by the shape of the violin) with statistical features (embedded box plots). The width of each violin reflects data density, with wider sections indicating higher concentration. The embedded box plot shows key statistics: the red horizontal line indicates the mean, the box spans the 25\% to 75\% data range, the black line in the middle represents the median. The vertical axis represents the metric values, with the different-colored boxes corresponding to the results obtained by the various methods. These plots allow for a clear comparison of distribution, central tendencies, and variability across methods, highlighting performance differences effectively.

We categorize the metrics into two groups reflecting fusion quality and dehazing quality, organized based on the dehazing methods used. Fig. \ref{fig6} presents violin plots of fusion metrics across three datasets, comparing our method with the combination of DIACMP for dehazing followed by fusion, and the Text-IF method. As shown in the $Q_{MI}$ metric’s violin plot, our method achieves the highest mean, with a more concentrated high-density distribution and relatively smaller variability, indicating that our data distribution is more centralized, demonstrating robust and stable performance in image fusion.
Fig. \ref{fig7} displays violin plots comparing our method with the combination of Dehazeformer for dehazing followed by fusion, and the Text-IF method. The results clearly show that our method exhibits superior fusion performance.

Fig. \ref{fig8} illustrates violin plots of dehazing metrics across three datasets, comparing our method with the combination of DIACMP for dehazing followed by fusion, and the Text-IF method. As depicted in the $Q_{NIQE}$  metric’s violin plot, our method achieves the lowest score while showing a more concentrated distribution skewed towards lower values and exhibiting smaller variability compared to other methods. Fig.  \ref{fig9} shows violin plots comparing our method with the combination of Dehazeformer for dehazing followed by fusion, and the Text-IF method. The violin plots for all three metrics demonstrate that our method achieves stable and outstanding performance in image restoration.

\begin{figure*}[ht!]
	\centering
	\includegraphics[width=1.0\textwidth]{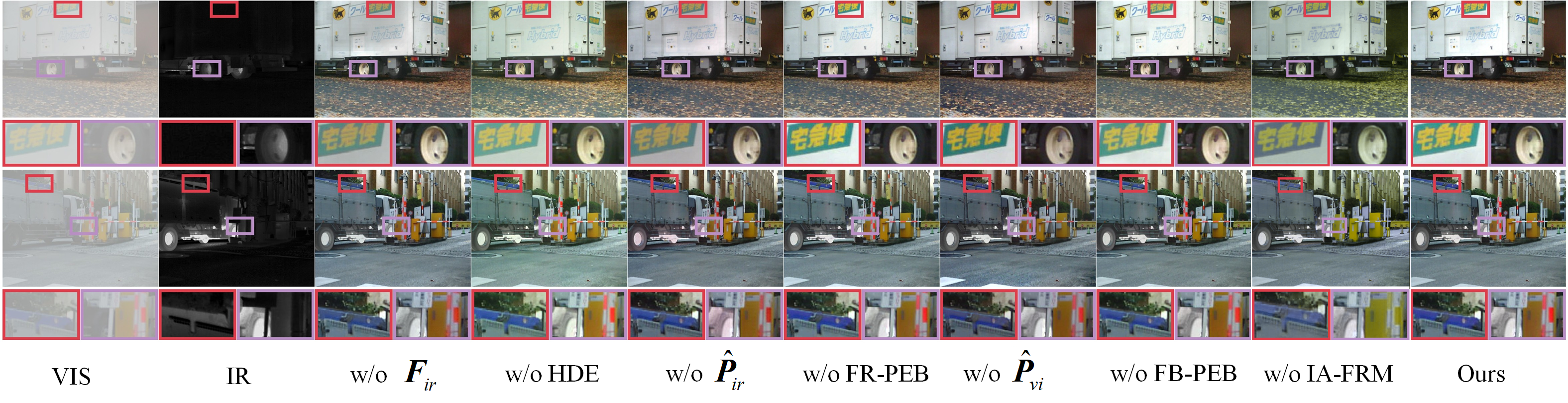}\vspace{-3mm}
	\caption{Ablation study on the fusion network design. The first two columns are the input source images, and the third to ninth columns are different fusion network.}\vspace{-3mm}
	\label{fig10}
\end{figure*}

\subsection{Ablation Study}
We design seven experimental settings to evaluate the effectiveness of each module. In the first setting, we remove the PGM-generated prompt ${\bm{\hat{P}}}_{ir}$, the haze density esitimation (HDE ) module, and the process of supplementing visible features with infrared features based on haze density, directly inputting ${\bm{F}}_{vi}$ into the Transformer Block for dehazing (denoted as ``w/o ${\bm{F}}_{ir}$") to assess the auxiliary role of infrared information. In the second setting, we replace the operation in Eq. (9) by directly adding ${\bm{\hat{F}}}_{ir}$ and ${\bm{F}}_{vi}$ (denoted as ``w/o HDE") to validate the impact of haze-density-based infrared integration. In the third setting, we remove the prompt embedding for ${\bm{\hat{P}}}_{ir}$ and the PEB module, injecting infrared features extracted by the encoder, guided by haze density ${\bm{H}}$, directly into ${\bm{F}}_{vi}$ (denoted as ``w/o ${\bm{\hat{P}}}_{ir}$") to evaluate the effect of prompt embedding. In the fourth setting, we test the effectiveness of the PEB module by removing it from IA-FRM, denoted as ``w/o FR-PEB". In the fifth setting, we remove ${\bm{\hat{P}}}_{vi}$ from MsPE-FM to verity the effectiveness of prompt embedding, denoted as ``w/o ${\bm{\hat{P}}}_{vi}$".In the sixth setting, we omit the PEB from the Fusion Block, performing feature concatenation and convolution directly (denoted as ``w/o FB-PEB") to validate its role in fusion. In the seventh setting, we remove the entire IA-FRM module (denoted as “w/o IA-FRM”) to evaluate the effectiveness of the dehazing component.

Table \ref{tab7} and Fig. \ref{fig10} demonstrate the impact of each module on fusion performance. While removing any module leads to performance decrease, these changes may not be readily visible in the qualitative results but are clearly reflected in the quantitative data. As shown in Table \ref{tab7}, when infrared information is not used to supplement visible features, the metrics ${Q_{PI}}$ and ${Q_{NIQE}}$ increase significantly, and all fusion metrics decrease, confirming the effectiveness of incorporating infrared information to enhance visible features. If the HDE module is excluded, and infrared information is directly injected into the visible image, model performance does not improve. Instead, color distortion appears in the fusion results, and objective evaluation metrics decline to varying degrees. Similar issues are observed in the setting ``w/o ${\bm{\hat{P}}}_{ir}$".  Additionally, for the settings ``w/o FR-PEB", `w/o ${\bm{\hat{P}}}_{vi}$" and ``w/o FB-PEB", the experimental results show slight degradation in detail retention, along with a decline in objective metrics, further validating the effectiveness of each module. In the “w/o IA-FRM” setting, when the IA-FRM module is removed and the hazy image pair is directly fed into the fusion network, it can be observed that the fusion results exhibit obvious color distortion and reduced contrast due to the presence of haze. Recognizable performance degradation is also reflected in the objective metrics, which further verifies the necessity of performing dehazing on visible images prior to fusion.

\begin{table}[t!]
	\centering	
	\caption{Quantitative results of seven ablation experiments on the MSRS dataset. The best and second-best performances are highlighted with \best{Red} and \second{Blue} backgrounds, respectively.}
	\renewcommand\arraystretch{1.2}
	{\footnotesize\centerline{\tabcolsep=8pt
			\begin{tabular}{cccccccccc}
				\hline
				\hline
				\textbf{Models} & ${Q}_{PI}\downarrow$ & ${Q}_{NIQE}\downarrow$ & ${Q}_{AB/F}\uparrow$ & ${Q}_{VIF}\uparrow$ & ${Q}_{SCD}\uparrow$ & ${Q}_{CV}\downarrow$ & ${Q}_{MI}\uparrow$ & ${Q}_{SF}\uparrow$ \\
				\hline
				w/o $\bm{F}_{ir}$ & 3.866 & 4.217 & \second{0.651} & \second{0.483} & 1.631 & 252.400 & 2.520 & \best{11.120} \\
				w/o HDE & 3.887 & 4.319 & 0.644 & 0.464 & 1.585 & 253.014 & 2.337 & 10.657 \\
				w/o $\bm{\hat{P}}_{ir}$ & 3.851 & 4.259 & 0.648 & 0.481 & 1.628 & 246.616 & 2.521 & 10.848 \\
				w/o FR-PEB & \second{3.819} & \best{4.099} & 0.644 & 0.475 & 1.601 & 251.357 & 2.349 & 10.942 \\
				w/o $\bm{\hat{P}}_{vi}$ & 3.820 & 4.212 & 0.638 & 0.478 & 1.598 & 294.239 & 2.132 & 10.902 \\
				w/o FB-PEB & 3.846 & 4.115 & 0.645 & 0.481 & \second{1.656} & \second{241.963} & \second{2.607} & 10.854 \\
				\textcolor{blue}{w/o IA-FRM} & 4.241 & 4.584 & 0.552 & 0.439 & 1.332 & 310.824 & 1.967 & 9.183 \\
				Ours & \best{3.811} & \second{4.110} & \best{0.652} & \best{0.490} & \best{1.662} & \best{238.420} & \best{2.720} & \second{11.050} \\
				\hline
	\end{tabular}}}
	\label{tab7}
\end{table}

\subsection{Hyperparameters Analysis}
Our method involves three key hyperparameters: the coefficient $\alpha$ for balancing the $L1$ loss, the number of Transformer blocks $L$ in IA-FRM, and the number of Fusion Blocks $M$ in MsPE-FM. During training, these hyperparameters are set to 2, 5, and 5, respectively. To validate the rationale behind these choices, we further conduct a detailed analysis of their impact on the model’s performance.

\textbf{The impact of $\alpha$ on model performance}. We fix $L$ and $M$ at 5 and study the impact of different values of $\alpha$ within the range (0, 5] on the model performance. Table \ref{tab8} presents the quantitative evaluation results of the fusion model on the MSRS dataset for various values of $\alpha$. The results indicate that when $\alpha$ is too small, the model performance does not reach its optimal level. Similarly, when $\alpha$ is too large, performance declines as well. The model achieves the best overall performance when $\alpha = 2$, which validates the choice of $\alpha = 2$ in this study. 

\begin{table}[!t]\footnotesize
	\centering{
		\caption{Quantitative analysis of six different $\alpha$ models on the MSRS dataset under the condition of $M=5$ and $L=5$. The best and second-best performances are highlighted with \best{Red} and \second{Blue} backgrounds, respectively.}
		\label{tab8}
		\renewcommand\arraystretch{1.2}
		{\footnotesize\centerline{\tabcolsep=8pt
				\begin{tabular}{ccccccccc}
					\hline
					\hline
					$\alpha$ & $Q_{MI}$ $\uparrow$ & $Q_{AB/F}$ $\uparrow$ & $Q_{CV}$ $\downarrow$ & $Q_{SCD}$ $\uparrow$ & $Q_{VIF}$ $\uparrow$ & $Q_{NIQE}$ $\downarrow$ & $Q_{PI}$ $\downarrow$ & $Q_{SF}$ $\uparrow$ \\ 
					\hline
					0.5 & 2.711 & 0.650 & 241.934 & 1.646 & \second{0.489} & 4.211 & \second{3.816} & \best{11.115} \\ 
					1 & \second{2.764} & \best{0.660} & \second{239.645} & 1.656 & 0.483 & 4.209 & 3.821 & \second{11.083} \\ 
					2 & 2.720 & 0.652 & \best{238.420} & \best{1.662} & \best{0.490} & \second{4.110} & \best{3.811} & 11.050 \\ 
					3 & 2.743 & \second{0.658} & 243.385 & 1.650 & 0.488 & \best{4.105} & 3.819 & 11.049 \\ 
					4 & \best{2.787} & 0.656 & 240.587 & \second{1.657} & \second{0.489} & 4.149 & 3.838 & 11.054 \\ 
					5 & 2.030 & 0.372 & 387.993 & 1.572 & 0.345 & 5.321 & 5.401 & 7.059 \\ 
					\hline
	\end{tabular}}}}
\end{table}

\textbf{The impact of $L$ on model performance}. To analyze the impact of different values of $L$ on the model performance, we fix $\alpha$ at 2 and $M$ at 5, and vary $L$ within the range [1, 7] to observe the corresponding performance changes. As shown in Table \ref{tab10}, the overall performance of the model improves as the number of Transformer Blocks increases. However, from the results for $L = 5$ and $L = 7$, it is evident that the performance gain has already plateaued, with only marginal improvements. Considering the increase in model parameters associated with larger $L$, we ultimately set $L$ to 5.

\begin{table}[!t]\footnotesize
	\centering {
		\caption{Quantitative analysis of four different $L$ models on the MSRS dataset under the condition of  $\alpha=2$ and $M=5$.  The best and second-best performances are highlighted with \best{Red} and \second{Blue} backgrounds, respectively.}
		\label{tab10}
		\renewcommand\arraystretch{1.2}
		{\footnotesize\centerline{\tabcolsep=8pt
				\begin{tabular}{ccccccccc}
					\hline
					\hline
					$L$ & $Q_{MI}$ $\uparrow$ & $Q_{AB/F}$ $\uparrow$ & $Q_{CV}$ $\downarrow$ & $Q_{SCD}$ $\uparrow$ & $Q_{VIF}$ $\uparrow$ & $Q_{NIQE}$ $\downarrow$ & $Q_{PI}$ $\downarrow$ & $Q_{SF}$ $\uparrow$ \\ 
					\hline
					1 & 2.353 & 0.592 & 273.282 & 1.492 & 0.411 & 4.526 & 4.113 & 9.988 \\ 
					3 & 2.592 & 0.636 & 259.921 & 1.589 & 0.468 & 4.391 & 3.981 & 10.147 \\ 
					5 & \second{2.720} & \second{0.652} & \best{238.420} & \best{1.662} & \second{0.490} & \second{4.110} & \best{3.811} & \best{11.050} \\ 
					7 & \best{2.769} & \best{0.659} & \second{240.167} & \second{1.641} & \best{0.493} & \best{4.082} & \second{3.826} & \second{10.992} \\ 
					\hline
	\end{tabular}}}}
\end{table}

\textbf{The impact of $M$ on model performance}. To analyze the impact of different values of $M$ on the model performance, we fix $\alpha$ at 2 and $L$ at 5, and vary $M$ within the range [1, 7] to observe the corresponding changes in performance. As shown in Table \ref{tab9}, the overall performance of the model improves as the number of Fusion Blocks increases. However, from the results for $M = 5$ and $M = 7$, it can be observed that the performance gain has already plateaued. Considering that further increasing the number of Fusion Blocks would significantly increase the number of model parameters, we ultimately set $M$ to 5.

\begin{table}[!t]\footnotesize
	\centering {
		\caption{Quantitative analysis of four different $M$ models on the MSRS dataset under the condition of  $\alpha=2$ and $L=5$. The best and second-best performances are highlighted with \best{Red} and \second{Blue} backgrounds, respectively.}
		\label{tab9}
		\renewcommand\arraystretch{1.2}
		{\footnotesize\centerline{\tabcolsep=8pt
				\begin{tabular}{ccccccccc}
					\hline
					\hline
					$M$ & $Q_{MI}$ $\uparrow$ & $Q_{AB/F}$ $\uparrow$ & $Q_{CV}$ $\downarrow$ & $Q_{SCD}$ $\uparrow$ & $Q_{VIF}$ $\uparrow$ & $Q_{NIQE}$ $\downarrow$ & $Q_{PI}$ $\downarrow$ & $Q_{SF}$ $\uparrow$ \\ 
					\hline
					1 & 2.567 & 0.651 & 251.174 & \best{1.669} & 0.484 & 4.113 & \second{3.816} & 10.967 \\ 
					3 & \best{2.915} & \best{0.662} & 244.317 & 1.643 & 0.488 & 4.146 & 3.839 & 10.982 \\ 
					5 & 2.720 & 0.652 & \best{238.420} & \second{1.662} & \second{0.490} & \second{4.110} & \best{3.811} & \best{11.050} \\ 
					7 & \second{2.832} & \second{0.659} & \second{238.883} & 1.656 & \best{0.491} & \best{4.108} & 3.840 & \second{11.027} \\ 
					\hline
	\end{tabular}}}}
\end{table}

\subsection{Complexity Analysis}
\begin{figure*}[ht!]
	\centering
	\includegraphics[width=0.8\textwidth]{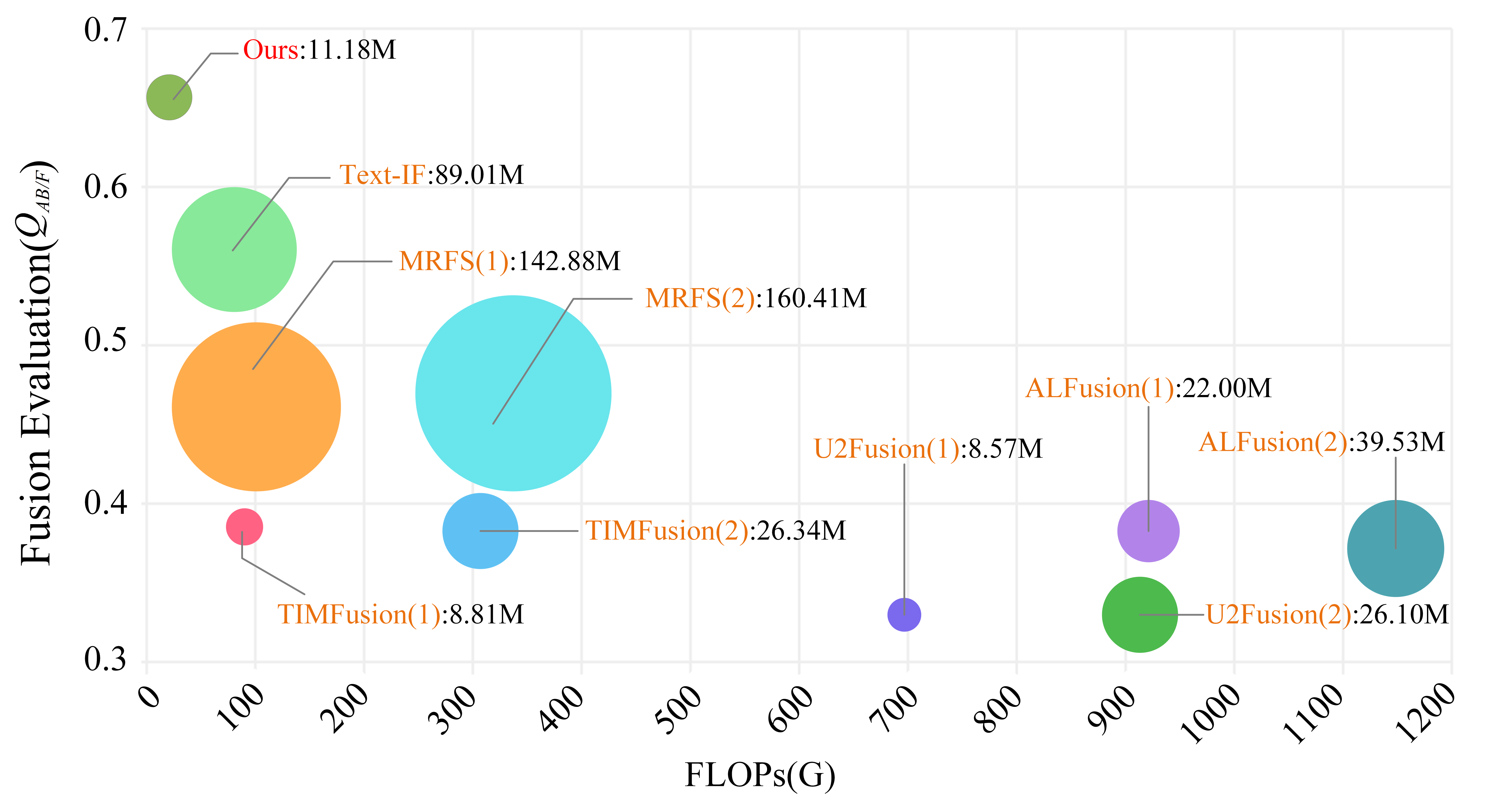}\vspace{-3mm}
	\caption{Model Complexity Analysis. In the figure, the x-axis represents the FLOPs(G) for models with input images of size  $256\times256$, the y-axis denotes the average value of the fusion metric, and the radius of the bubbles reflects the model's parameter count. In the method names, (1) refers to the result obtained by first applying DIACMP for dehazing followed by fusion, and (2) refers to the result obtained by first applying Dehazeformer for dehazing followed by fusion.}\vspace{-3mm}
	\label{fig11}
\end{figure*}

This paper employs a single-stage framework for hazy image fusion, which significantly reduces the model's complexity and parameter count. To validate this advantage, we test the FLOPs and parameter count of each model and plot a bubble chart, where the bubble size denotes the fusion metric $Q_{AB/F}$ computed on the MSRS dataset. As shown in Fig. \ref{fig11}, our model achieves optimal performance while maintaining a low parameter count and the lowest computational complexity, fully demonstrating its efficiency and suitability for practical deployment.

\subsection{Limitations and Future Work}
Although our method effectively handles hazy image fusion with high quality, real-world scenarios may still present challenging weather conditions such as low light, snow, and rain. Our approach does not currently account for the impact of these factors. While some existing methods have attempted to address this issue, they fail to effectively balance the impact of different degradations on the fusion results. Furthermore, these methods typically assume that the degradation types in the images have been encountered by the model, and the model performance tends to be suboptimal in scenarios with unseen degradations. Therefore, future work will focus on designing a multi-degradation joint processing framework to enable the model to effectively perform fusion and restoration even in scenarios with unseen degradations.

\section{Conclusion}
This paper presents an infrared-assisted joint learning framework for IR-VIS image fusion under hazy conditions. By integrating dehazing and fusion tasks into a single-stage framework with collaborative training, our method effectively enhances feature restoration and fusion performance. Experimental results demonstrate that our approach produces clear, haze-free fusion images, outperforming traditional two-stage methods and existing multi-task fusion frameworks. The lightweight and compact model structure also ensures practical deployment, making it a valuable solution for hazy image restoration and fusion.

\section*{Acknowledgments}
This work was supported in part by the National Natural Science Foundation of China under Grants (62161015, 62276120), 
and in part by the Yunnan Fundamental Research Projects under Grants (202301AV070004, 202501AS070123, 202401AS070640).

\bibliography{mybibfile}

@String(CVPR= {IEEE Conf. Comput. Vis. Pattern Recog.})

@String(ICLR = {Int. Conf. Learn. Represent.})

@String(IJCAI = {IJCAI})

@String(CVPR  = {CVPR})

@String(ICLR  = {ICLR})

@article{1,
	author =       {Huafeng Li and Yitang Wang and Zhao Yang and Ruxin Wang and Xiang Li and Dapeng Tao},
	title =        {Discriminative Dictionary Learning Based Multiple Component Decomposition for Detail-Preserving Noisy Image Fusion},
	journal =      {IEEE Transactions on Instrumentation and Measurement},
	year =         {2020},
	number={4},
	volume =       {69},
	pages =        {1082--1102},
}

@article{2,
	author = {Huafeng Li and Junzhi Zhao and Jinxing Li and Zhengtao Yu and Guangming Lu},
	title = {Feature dynamic alignment and refinement for infrared–visible image fusion: Translation robust fusion},
	journal = {Information Fusion},
	volume = {95},
	pages = {26--41},
	year = {2023},
}

@article{3,
	author = {Huafeng Li and Junyu Liu and Yafei Zhang and Yu Liu},
	title = {A Deep Learning Framework for Infrared and Visible Image Fusion
	Without Strict Registration},
	journal = {International Journal of Computer Vision},
	year = {2024},
	volume={132}, 
	pages={1625–1644},
}

@article{4,
	title={Coconet: Coupled contrastive learning network with multi-level feature ensemble for multi-modality image fusion},
	author={Liu, Jinyuan and Lin, Runjia and Wu, Guanyao and Liu, Risheng and Luo, Zhongxuan and Fan, Xin},
	journal={International Journal of Computer Vision},
	volume={132},
	number={5},
	pages={1748--1775},
	year={2024},
	publisher={Springer}
}

@article{5,
	author={Li, Hui and Xu, Tianyang and Wu, Xiao-Jun and Lu, Jiwen and Kittler, Josef},
	title={LRRNet: A Novel Representation Learning Guided Fusion Network for Infrared and Visible Images}, 
	journal={IEEE Transactions on Pattern Analysis and Machine Intelligence}, 
	year={2023},
	volume={45},
	number={9},
	pages={11040--11052},
	publisher={IEEE}
}

@ARTICLE{11,
	author={Li, Jing and Huo, Hongtao and Li, Chang and Wang, Renhua and Feng, Qi},
	journal={IEEE Transactions on Multimedia}, 
	title={AttentionFGAN: Infrared and Visible Image Fusion Using Attention-Based Generative Adversarial Networks}, 
	year={2021},
	volume={23},
	number={},
	pages={1383--1396},
	publisher={IEEE}
}

@article{12,
	author = {Jiayi Ma and Wei Yu and Pengwei Liang and Chang Li and Junjun Jiang},
	journal = {Information Fusion},
	title = {FusionGAN: A generative adversarial network for infrared and visible image fusion},
	year = {2019},
	volume = {48},
	pages = {11--26},
	publisher={Elsevier}
}

@ARTICLE{13,
	author={Liu, Jinyuan and Fan, Xin and Jiang, Ji and Liu, Risheng and Luo, Zhongxuan},
	journal={IEEE Transactions on Circuits and Systems for Video Technology}, 
	title={Learning a Deep Multi-Scale Feature Ensemble and an Edge-Attention Guidance for Image Fusion}, 
	year={2022},
	volume={32},
	number={1},
	pages={105--119},
	publisher={IEEE}
}

@article{14,
	author={Li, Hui and Wu, Xiao-Jun},
	journal={IEEE Transactions on Image Processing}, 
	title={DenseFuse: A Fusion Approach to Infrared and Visible Images}, 
	year={2019},
	volume={28},
	number={5},
	pages={2614--2623},
	publisher={IEEE}
}

@InProceedings{16,
	author = {Liu, Jinyuan and Fan, Xin and Huang, Zhanbo and Wu, Guanyao and Liu, Risheng and Zhong, Wei and Luo, Zhongxuan},
	title  = {Target-Aware Dual Adversarial Learning and a Multi-Scenario Multi-Modality Benchmark To Fuse Infrared and Visible for Object Detection},
	booktitle = {Proceedings of the IEEE/CVF Conference on Computer Vision and Pattern Recognition (CVPR)},
	year = {2022},
	pages = {5802-5811}
}

@InProceedings{17,
	author    = {Zhao, Wenda and Xie, Shigeng and Zhao, Fan and He, You and Lu, Huchuan},
	title     = {MetaFusion: Infrared and Visible Image Fusion via Meta-Feature Embedding From Object Detection},
	booktitle = {Proceedings of the IEEE/CVF Conference on Computer Vision and Pattern Recognition (CVPR)},
	year      = {2023},
	pages     = {13955-13965}
}

@inproceedings{18,
	title={Learning a Generative Model for Fusing Infrared and Visible Images via Conditional Generative Adversarial Network with Dual Discriminators},
	author={Xu, Han and Liang, Pengwei and Yu, Wei and Jiang, Junjun and Ma, Jiayi},
	booktitle={Proceedings of the Twenty-Eighth International Joint Conference on Artificial Intelligence (IJCAI)},
	pages={3954--3960},
	year={2019}
}

@article{19,
	title={DDcGAN: A dual-discriminator conditional generative adversarial network for multi-resolution image fusion},
	author={Ma, Jiayi and Xu, Han and Jiang, Junjun and Mei, Xiaoguang and Zhang, Xiao-Ping},
	journal={IEEE Transactions on Image Processing},
	volume={29},
	pages={4980--4995},
	year={2020},
	publisher={IEEE}
}

@article{20,
	title={AFT: Adaptive fusion transformer for visible and infrared images},
	author={Chang, Zhihao and Feng, Zhixi and Yang, Shuyuan and Gao, Quanwei},
	journal={IEEE Transactions on Image Processing},
	volume={32},
	pages={2077--2092},
	year={2023},
	publisher={IEEE}
}

@article{21,
	title={YDTR: Infrared and visible image fusion via Y-shape dynamic transformer},
	author={Tang, Wei and He, Fazhi and Liu, Yu},
	journal={IEEE Transactions on Multimedia},
	volume={25},
	pages={5413--5428},
	year={2022},
	publisher={IEEE}
}

@article{22,
	title={HitFusion: Infrared and Visible Image Fusion for High-level Vision Tasks Using Transformer},
	author={Chen, Jun and Ding, Jianfeng and Ma, Jiayi},
	journal={IEEE Transactions on Multimedia},
	volume={26},
	pages={10145--10159},
	year={2024},
	publisher={IEEE}
}

@article{24,
	title={DIVFusion: Darkness-free infrared and visible image fusion},
	author={Tang, Linfeng and Xiang, Xinyu and Zhang, Hao and Gong, Meiqi and Ma, Jiayi},
	journal={Information Fusion},
	volume={91},
	pages={477--493},
	year={2023},
	publisher={Elsevier}
}

@article{25,
	title={IAIFNet: An Illumination-Aware Infrared and Visible Image Fusion Network},
	author={Yang, Qiao and Zhang, Yu and Zhao, Zijing and Zhang, Jian and Zhang, Shunli},
	journal={IEEE Signal Processing Letters},
	volume={31},
	pages={1374--1378},
	year={2024},
	publisher={IEEE}
}

@article{26,
	title={LENFusion: A Joint Low-Light Enhancement and Fusion Network for Nighttime Infrared and Visible Image Fusion},
	author={Chen, Jun and Yang, Liling and Liu, Wei and Tian, Xin and Ma, Jiayi},
	journal={IEEE Transactions on Instrumentation and Measurement},
	volume={73},
	pages={1--15},
	year={2024},
	publisher={IEEE}
}

@article{27,
	title={Heterogeneous knowledge distillation for simultaneous infrared-visible image fusion and super-resolution},
	author={Xiao, Wanxin and Zhang, Yafei and Wang, Hongbin and Li, Fan and Jin, Hua},
	journal={IEEE Transactions on Instrumentation and Measurement},
	volume={71},
	pages={1--15},
	year={2022},
	publisher={IEEE}
}

@article{28,
	author =       {Huafeng Li and Yueliang Cen and Yu Liu and Xun Chen and Zhengtao Yu},
	title =        {Different Input Resolutions and Arbitrary Output Resolution: A Meta Learning-Based Deep Framework for Infrared and Visible Image Fusion},
	journal =      {IEEE Transactions on Image Processing},
	year =         {2021},
	volume = {30},
	pages= {4070--4083},
}

@article{29,
	title={Decomposition based and Interference Perception for Infrared and Visible Image Fusion in Complex Scenes},
	author={Li, Xilai and Li, Xiaosong and Tan, Haishu},
	journal={arXiv preprint arXiv: 2402. 02096},
	year={2024}
}

@article{30,
	title={Physical Perception Network and an All-weather Multi-modality Benchmark for Adverse Weather Image Fusion},
	author={Li, Xilai and Liu, Wuyang and Li, Xiaosong and Tan, Haishu},
	journal={arXiv preprint arXiv: 2402. 02090},
	year={2024}
}

@inproceedings{31,
	title={Text-IF: Leveraging Semantic Text Guidance for Degradation-Aware and Interactive Image Fusion},
	author={Yi, Xunpeng and Xu, Han and Zhang, Hao and Tang, Linfeng and Ma, Jiayi},
	booktitle={Proceedings of the IEEE/CVF Conference on Computer Vision and Pattern Recognition (CVPR)},
	pages={27026--27035},
	year={2024}
}

@article{32,
	title={VIFNet: An end-to-end visible-infrared fusion network for image dehazing},
	author={Yu, Meng and Cui, Te and Lu, Haoyang and Yue, Yufeng},
	journal={Neurocomputing},
	pages={128105},
	year={2024},
	publisher={Elsevier}
}

@INPROCEEDINGS{33,
	author={Zamir, Syed Waqas and Arora, Aditya and Khan, Salman and Hayat, Munawar and Khan, Fahad Shahbaz and Yang, Ming–Hsuan},
	booktitle={2022 IEEE/CVF Conference on Computer Vision and Pattern Recognition (CVPR)}, 
	title={Restormer: Efficient Transformer for High-Resolution Image Restoration}, 
	year={2022},
	pages={5718-5729}
}

@article{34,
	title={Single image haze removal using dark channel prior},
	author={He, Kaiming and Sun, Jian and Tang, Xiaoou},
	journal={IEEE Transactions on Pattern Analysis and Machine Intelligence},
	volume={33},
	number={12},
	pages={2341--2353},
	year={2011},
	publisher={IEEE}
}

@ARTICLE{35,
	author={Yue, Jun and Fang, Leyuan and Xia, Shaobo and Deng, Yue and Ma, Jiayi},
	journal={IEEE Transactions on Image Processing}, 
	title={Dif-Fusion: Toward High Color Fidelity in Infrared and Visible Image Fusion With Diffusion Models}, 
	year={2023},
	volume={32},
	pages={5705-5720},
}

@article{36,
	title={PIAFusion: A progressive infrared and visible image fusion network based on illumination aware},
	author={Tang, Linfeng and Yuan, Jiteng and Zhang, Hao and Jiang, Xingyu and Ma, Jiayi},
	journal={Information Fusion},
	volume = {83-84},
	pages = {79--92},
	year = {2022}
}

@ARTICLE{38,
	author={Xu, Han and Ma, Jiayi and Jiang, Junjun and Guo, Xiaojie and Ling, Haibin},
	journal={IEEE Transactions on Pattern Analysis and Machine Intelligence}, 
	title={U2Fusion: A Unified Unsupervised Image Fusion Network}, 
	year={2022},
	volume={44},
	number={1},
	pages={502-518},
	publisher={IEEE}
}

@inproceedings{39,
	title={Adam: A method for stochastic optimization},
	author={Kingma, Diederik P. and Ba, Jimmy},
	booktitle =  {International Conference on Learning Representations (ICLR)},
	year = {2015},
}

@article{40,
	title={Information measure for performance of image fusion},
	author={Qu, Guihong and Zhang, Dali and Yan, Pingfan},
	journal={Electronics Letters},
	volume={38},
	number={7},
	pages={1},
	year={2002},
}

@article{41,
	title={Objective image fusion performance measure},
	author={Xydeas, Costas S and Petrovic, Vladimir and others},
	journal={Electronics Letters},
	volume={36},
	number={4},
	pages={308--309},
	year={2000},
}

@article{42,
	title={A human perception inspired quality metric for image fusion based on regional information},
	author={Chen, Hao and Varshney, Pramod K},
	journal={Information Fusion},
	volume={8},
	number={2},
	pages={193--207},
	year={2007},
	publisher={Elsevier}
}

@article{43,
	title={A new image quality metric for image fusion: The sum of the correlations of differences},
	author={Aslantas, V and Bendes, Emre},
	journal={Aeu-International Journal of Electronics and Communications},
	volume={69},
	number={12},
	pages={1890--1896},
	year={2015},
	publisher={Elsevier}
}

@article{44,
	title={A new image fusion performance metric based on visual information fidelity},
	author={Han, Yu and Cai, Yunze and Cao, Yin and Xu, Xiaoming},
	journal={Information Fusion},
	volume={14},
	number={2},
	pages={127--135},
	year={2013},
	publisher={Elsevier}
}

@inproceedings{45,
	title={The 2018 PIRM challenge on perceptual image super-resolution},
	author={Blau, Yochai and Mechrez, Roey and Timofte, Radu and Michaeli, Tomer and Zelnik-Manor, Lihi},
	booktitle={Proceedings of the European Conference on Computer Vision Workshops (ECCVW)},
	pages={334--355},
	year={2018}
}

@article{46,
	title={Making a ``completely blind'' image quality analyzer},
	author={Mittal, Anish and Soundararajan, Rajiv and Bovik, Alan C.},
	journal={IEEE Signal Processing Letters},
	volume={20},
	number={3},
	pages={209--212},
	year={2013},
	publisher={IEEE}
}

@article{47,
	title={Image quality measures and their performance},
	author={Eskicioglu, Ahmet M and Fisher, Paul S},
	journal={IEEE Transactions on Communications},
	volume={43},
	number={12},
	pages={2959--2965},
	year={1995},
	publisher={IEEE}
}

@inproceedings{48,
	title={Depth Information Assisted Collaborative Mutual Promotion Network for Single Image Dehazing},
	author={Zhang, Yafei and Zhou, Shen and Li, Huafeng},
	booktitle={Proceedings of the IEEE/CVF Conference on Computer Vision and Pattern Recognition (CVPR)},
	pages={2846--2855},
	year={2024}
}

@article{49,
	title={Vision transformers for single image dehazing},
	author={Song, Yuda and He, Zhuqing and Qian, Hui and Du, Xin},
	journal={IEEE Transactions on Image Processing},
	volume={32},
	pages={1927--1941},
	year={2023},
	publisher={IEEE}
}

@article{50,
	title={A task-guided, implicitly-searched and metainitialized deep model for image fusion},
	author={Liu, Risheng and Liu, Zhu and Liu, Jinyuan and Fan, Xin and Luo, Zhongxuan},
	journal={IEEE Transactions on Pattern Analysis and Machine Intelligence},
	volume={46},
	number={10},
	pages={6594--6609},
	year={2024},
	publisher={IEEE}
}

@inproceedings{51,
	title={MRFS: Mutually Reinforcing Image Fusion and Segmentation},
	author={Zhang, Hao and Zuo, Xuhui and Jiang, Jie and Guo, Chunchao and Ma, Jiayi},
	booktitle={Proceedings of the IEEE/CVF Conference on Computer Vision and Pattern Recognition (CVPR)},
	pages={26974--26983},
	year={2024}
}

@ARTICLE{54,
	title={ACFNet: An adaptive cross-fusion network for infrared and visible image fusion},
	author={Chen, Xiaoxuan and Xu, Shuwen and Hu, Shaohai and Ma, Xiaole},
	journal={Pattern Recognition},
	volume={159},
	pages={111098},
	year={2025},
	publisher={Elsevier}
}

@article{57,
	title={Quaternion nuclear norm minus frobenius norm minimization for color image reconstruction},
	author={Guo, Yu and Chen, Guoqing and Zeng, Tieyong and Jin, Qiyu and Ng, Michael Kwok-Po},
	journal={Pattern Recognition},
	volume={158},
	pages={110986},
	year={2025},
	publisher={Elsevier}
}

@article{58,
	title={Investigating intrinsic degradation factors by multi-branch aggregation for real-world underwater image enhancement},
	author={Xue, Xinwei and Li, Zexuan and Ma, Long and Jia, Qi and Liu, Risheng and Fan, Xin},
	journal={Pattern recognition},
	volume={133},
	pages={109041},
	year={2023},
	publisher={Elsevier}
}

@article{60,
	title={Dc-net: Divide-and-conquer for salient object detection},
	author={Zhu, Jiayi and Qin, Xuebin and Elsaddik, Abdulmotaleb},
	journal={Pattern Recognition},
	volume={157},
	pages={110903},
	year={2025},
	publisher={Elsevier}
}

@article{62,
	title={HDR reconstruction from a single exposure LDR using texture and structure dual-stream generation},
	author={Chen, Yu-Hsiang and Ruan, Shanq-Jang},
	journal={Pattern Recognition},
	volume={159},
	pages={111127},
	year={2025},
	publisher={Elsevier}
}

@inproceedings{DenoIF,
title={Deno-{IF}: Unsupervised Noisy Visible and Infrared Image Fusion Method},
author={Han Xu and Yuyang Li and Yunfei Deng and Jiayi Ma and Guangcan Liu},
booktitle={The Thirty-ninth Annual Conference on Neural Information Processing Systems},
year={2025},
url={https://openreview.net/forum?id=36cKp4tsHF}
}

@article{OmniFuse,
  title={OmniFuse: Composite Degradation-Robust Image Fusion with Language-Driven Semantics},
  author={Zhang, Hao and Cao, Lei and Zuo, Xuhui and Shao, Zhenfeng and Ma, Jiayi},
  journal={IEEE Transactions on Pattern Analysis and Machine Intelligence},
  year={2025},
  publisher={IEEE}
}

@article{TextDiFuse,
  title={Text-DiFuse: An interactive multi-modal image fusion framework based on text-modulated diffusion model},
  author={Zhang, Hao and Cao, Lei and Ma, Jiayi},
  journal={Advances in Neural Information Processing Systems},
  volume={37},
  pages={39552--39572},
  year={2024}
}

@ARTICLE{freefusion,
  author={Zhao, Wenda and Cui, Hengshuai and Wang, Haipeng and He, You and Lu, Huchuan},
  journal={IEEE Transactions on Pattern Analysis and Machine Intelligence}, 
  title={FreeFusion: Infrared and Visible Image Fusion via Cross Reconstruction Learning}, 
  year={2025},
  volume={47},
  number={9},
  pages={8040-8056},
  doi={10.1109/TPAMI.2025.3572599}}

@INPROCEEDINGS{ship,
  author={Zheng, Naishan and Zhou, Man and Huang, Jie and Hou, Junming and Li, Haoying and Xu, Yuan and Zhao, Feng},
  booktitle={2024 IEEE/CVF Conference on Computer Vision and Pattern Recognition (CVPR)}, 
  title={Probing Synergistic High-Order Interaction in Infrared and Visible Image Fusion}, 
  year={2024},
  volume={},
  number={},
  pages={26374-26385},
  doi={10.1109/CVPR52733.2024.02492}}

@ARTICLE{cdtfusion,
  author={Zhao, Wenda and Wang, Wenbo and Wang, Haipeng and He, You and Lu, Huchuan},
  journal={IEEE Transactions on Pattern Analysis and Machine Intelligence}, 
  title={CDTFusion: Crossing Domain and Task for Infrared and Visible Image Fusion}, 
  year={2025},
  volume={},
  number={},
  pages={1-14},
  doi={10.1109/TPAMI.2025.3614704}}
\end{document}